\documentclass[10pt,twocolumn,letterpaper]{article}
\usepackage[pagenumbers]{wacv} 

\usepackage{graphicx,verbatim}
\usepackage{amsmath}
\usepackage{booktabs}
\usepackage{multirow}
\usepackage{float}
\usepackage{tabularx}
\usepackage{array}
\usepackage{tcolorbox}
\usepackage[normalem]{ulem}
\usepackage{placeins}
\definecolor{wacvblue}{rgb}{0.21,0.49,0.74}
\usepackage[pagebackref,breaklinks,colorlinks,allcolors=wacvblue]{hyperref}

\def\wacvPaperID{1334} 
\def\confName{WACV}
\def\confYear{2027}

\usepackage{soul}
\newcommand{\sig}{\textsuperscript{\(\dagger\)}}
\usepackage[accsupp]{axessibility}  

\usepackage{xcolor}
\title{SMILE: A Structured Multi-Agent Clinical Reasoning for Liver Ultrasound Understanding}

\author{
Haris Ghafoor$^{\dagger}$\\
Yonsei University\\
\and
Minjee Seo$^{\dagger}$\\
Yonsei University\\
\and
Seonaeng Cho\\
Yonsei University\\
\and
Keungmo Yang\\
The Catholic University of Korea\\
\and
Jaejun Lee\\
The Catholic University of Korea\\
\and
Kyungho Yoon$^{\star}$\\
Yonsei University\\
}

\begin{document}
\AddToHookNext{shipout/foreground}{%
  \put(55,-25){%
    \begingroup
    \setlength{\fboxsep}{5pt}%
    \colorbox{black!7}{%
      \parbox{\dimexpr\textwidth-2\fboxsep\relax}{%
        \raggedright
        \fontsize{8}{8.5}\selectfont
        \copyright\ 2027 IEEE. Personal use of this material is permitted. Permission from IEEE must be obtained
        for all other uses, in any current or future media,
        including reprinting/republishing this material for
        advertising or promotional purposes, creating new
        collective works, for resale or redistribution to
        servers or lists, or reuse of any copyrighted
        component of this work in other works. Accepted for publication in the Proceedings of the 2027 IEEE/CVF Winter Conference on Applications of Computer Vision (WACV 2027), Disney Springs, Buena Vista, Florida, USA, January 4--8, 2027.
      }%
    }%
    \endgroup
  }%
}

\maketitle

%

\begingroup
\renewcommand{\thefootnote}{}
\footnotetext{%
\textsuperscript{\ensuremath{\dagger}} Equal Contribution.
\hspace{0.5em}
\textsuperscript{\ensuremath{\star}} Corresponding Author.
}
\endgroup

\begin{abstract}
Automated liver ultrasound report generation improves the diagnosis of liver diseases by producing clinically structured descriptions of anatomical and disease-related findings.
However, current generative vision-language models (VLMs) often rely on a single reasoning pathway that maps ultrasound appearance directly to textual descriptions.
In addition to the inherent vision-language modality gap, this design is limited for liver ultrasound because disease interpretation is constrained by subtle visual evidence, heterogeneous disease patterns, and limited paired supervision.
To address these limitations, we propose SMILE, a Structured Multi-agent clInicaL rEasoning framework that uses multiple complementary specialist agents to interpret ultrasound images and consolidates their outputs through a judge-guided consensus mechanism.
Our approach employs three specialized agents: an image interpretation agent for morphological descriptions, a clinical-stage reasoning agent, and a visual question answering (VQA) agent for targeted queries.
These agent responses are then aggregated and scrutinized by a judge VLM using a consensus filtering prompt to produce a unified, clinically coherent output.
Across a public liver ultrasound dataset and an internal clinical dataset, SMILE consistently outperforms state-of-the-art VLMs, achieving up to 19.0\% improvement in ROUGE-1 and 18.8\% improvement in clinical F1-score, while reaching VQA F1-scores of 0.82 and 0.65 on public and internal datasets, respectively.
These results demonstrate the potential of SMILE for data-scarce and limited-supervision clinical settings.
Our models, datasets, and code are available \href{https://github.com/CMME-Lab/SMILE}{here}.
\end{abstract}
\begin{figure}[!ht]
\centering
\includegraphics[width=0.95\columnwidth]{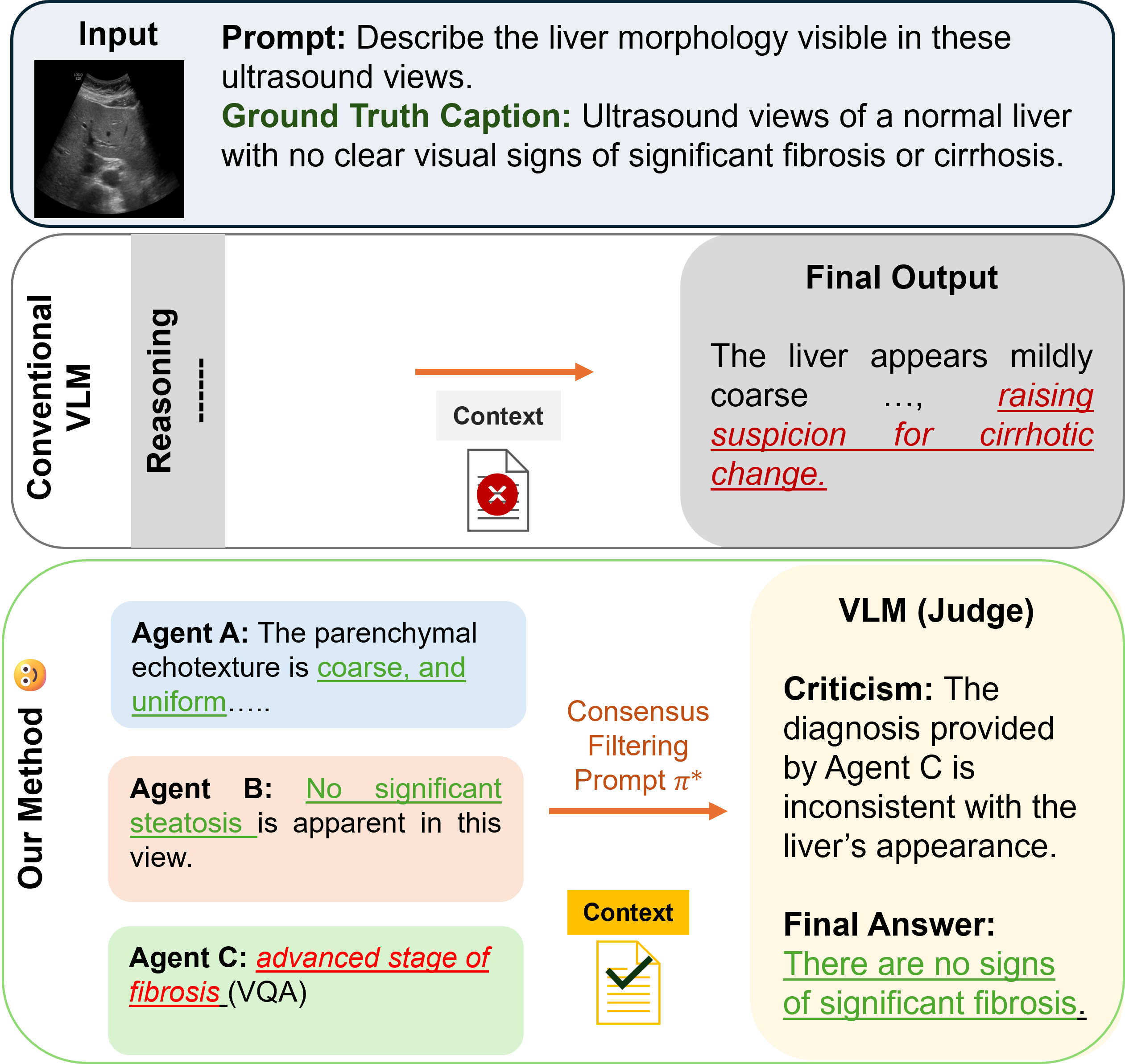}
\caption{Comparison with existing VLMs.
Top (Conventional VLM): Utilizing a single VLM, pre-trained on a single source of clinical information, which generates responses with hallucinated findings.
Bottom (Our Method): SMILE uses complementary specialist agents adapted with radiology-aligned and pathology-aligned supervision to extract complementary clinical evidence from liver ultrasound images.   
Red text indicates hallucinated findings, and green text indicates findings corrected by SMILE.
}
\label{fig:teaser}
\end{figure}


\section{Introduction}
Ultrasound imaging plays a central role in the evaluation of liver diseases due to its accessibility, real-time feedback, cost-effectiveness, and non-invasive assessment \cite{friedrich2009liver}.
However, reliable diagnosis of liver diseases remains challenging because liver pathology presents heterogeneous and often ill-defined sonographic patterns, and ultrasound images are affected by data-acquisition artifacts such as speckle noise and acoustic shadowing \cite{2002.21.9.1023,d2019b}. As a representative example, staging liver fibrosis from conventional ultrasound is particularly difficult because early fibrotic changes may not produce overt structural abnormalities \cite{bataller2005liver,wynn2012mechanisms}.

Recent advances in artificial intelligence (AI) have enabled automated analysis of ultrasound images \cite{lee2020deep,joo2023classification}. In particular, large language models (LLMs) have recently been explored for medical image understanding due to their strong capabilities in semantic abstraction and structured reasoning \cite{yang2025application,liu2024claude,hirosawa2023chatgpt}.
Unlike purely discriminative vision models, the integration of LLMs with vision systems can externalize intermediate reasoning steps through natural language, providing traceable and clinically interpretable explanations \cite{chen-etal-2024-towards-injecting}.
However, in liver ultrasound analysis, the primary challenge extends beyond text generation itself.
The visual evidence provided by ultrasound is often under-constrained, allowing multiple clinically plausible interpretations to coexist for a single image.
Conventional fine-tuning of single-agent LLMs implicitly assumes that a solitary reasoning trajectory is sufficient to resolve this ambiguity \cite{kim2025medical}.
Under limited supervision and small-data regimes, this assumption leads to unstable reasoning behavior, loss of context awareness, and hallucinated findings that are weakly grounded in the image \cite{alansari2025large,wang2024_loss_of_context_awareness,zhuo2024prosa}.

To improve the reasoning capabilities of LLMs, several studies have explored collaboration among multiple large language models for complex medical reasoning tasks \cite{tang2024medagents,chen2024reconcile,kim2024mdagents,tran2025multiagent_collabortion,zhang2025_multiagent}.
These methods improve the reasoning of a LLM through role-playing, where each agent assumes a distinct expert perspective and proposes diverse plans that capture different aspects of the clinical problem.
However, the benefit of such reasoning models depends strongly on the design of the collaboration mechanism among LLMs \cite{tran2025multiagent_collabortion}.
Moreover, without explicit constraints that enforce complementary reasoning, multi-agent systems may generate homogeneous, low-quality responses and induce misconceptions \cite{becker2026stay,estornell2024multiagent_issues}.
These challenges are further highlighted in clinical imaging tasks, where limited access to independent and modality-specific evidence can reduce agent diversity and amplify unsupported clinical interpretations.

To address these limitations, we propose SMILE, a structured multi-agent clinical reasoning framework for liver ultrasound image analysis.
Rather than depending on a single reasoning pathway, our approach decomposes the diagnostic process into complementary agents: (1) an image interpretation agent that converts visual features into structured morphological descriptions, (2) a clinical-stage reasoning agent guided by pathology-aligned supervision, and (3) a visual question answering (VQA) agent that improves diagnostic performance through targeted queries.
These agentic responses are then evaluated by a dedicated LLM, referred to as the judge, that improves the consistency and clinical plausibility of the agents' responses and synthesizes them into a unified, diagnostically coherent output (Fig.~\ref{fig:teaser}).
Unlike existing multi-LLM medical reasoning frameworks, which operate on a specific clinical modality and require resource-intensive post-training, we only employ lightweight agents that can be adapted through simple fine-tuning with limited labeled datasets, making the framework suitable for data-scarce scenarios.

Our contributions can be summarized as follows: 1) We formulate liver ultrasound understanding as an under-constrained clinical reasoning problem; 2) We propose a structured multi-agent framework that stabilizes reasoning through cross-agent consistency; 3) We demonstrate improved robustness and interpretability by assessing medical image understanding under limited data conditions.


\begin{figure*}[!t]
\centering
\includegraphics[
width=\textwidth,
height=0.35\textwidth,
keepaspectratio
]
{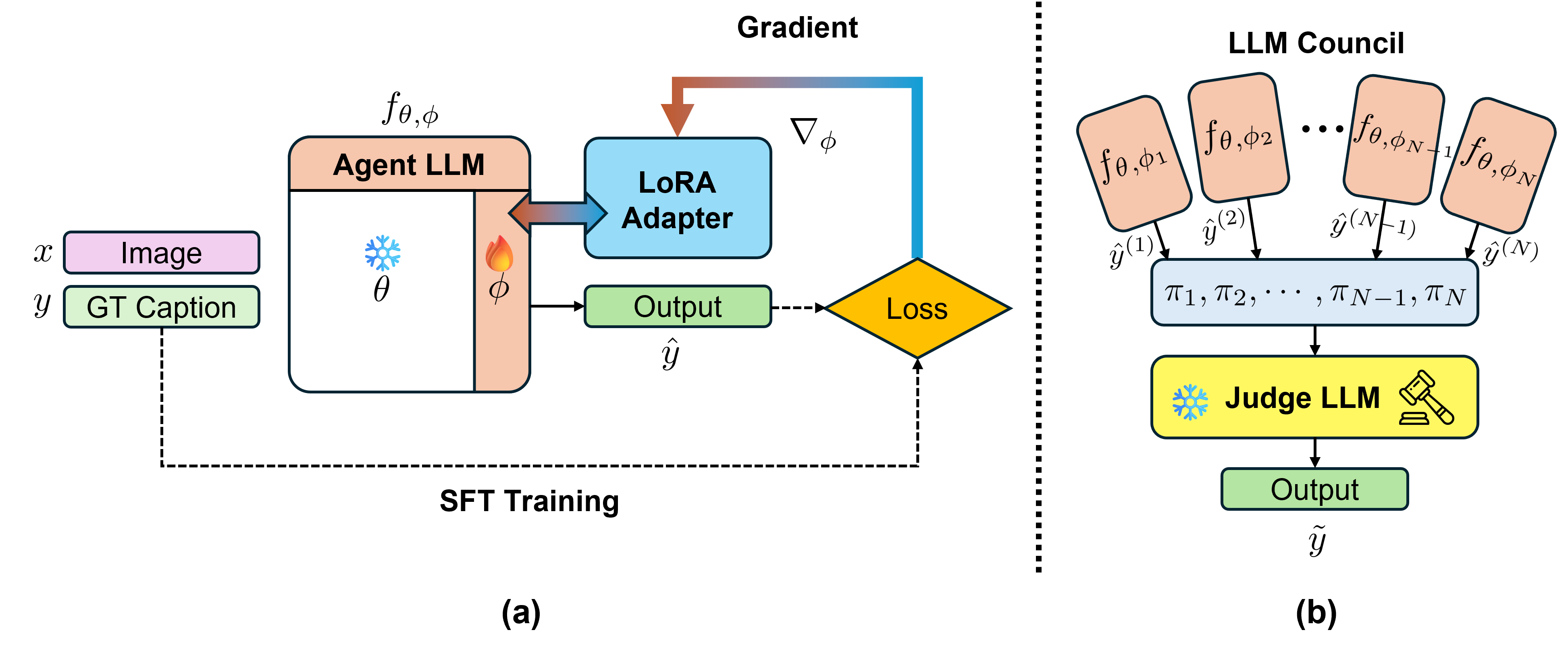}
\caption{An overview of our proposed method. (a) Training procedure of a single agent as explained in Section \ref{subsec:agent_training}. (b) Inference from multi-agent under consensus filtering by a judge to obtain a high-fidelity liver ultrasound report.}
\label{fig:method}
\end{figure*}

\section{Related Work}
\label{sec:related_work}
\subsection{Vision Language Models in Medical Report Generation}
Medical report generation has evolved from task-specific image-to-text generation toward generative vision-language models (VLMs) with broader clinical reasoning ability.
Early methods such as R2Gen used memory-driven Transformer architectures to generate long clinical reports directly from medical images \cite{chen2020r2gen_firsts}.
Subsequent report generation methods addressed the cross-modal semantic gap between visual findings and clinical language, aiming to reduce inconsistent or visually unsupported descriptions \cite{wang2022medical,li2022cross_issues_mrg, wang2023r2gengpt}.
With the rise of large language models in natural text generation, recent VLMs have moved beyond task-specific decoders by connecting visual encoders with pretrained language models for open-ended multi-modal generation.
Flamingo and Med-Flamingo introduced few-shot multi-modal learning through visual resampling and gated cross-attention \cite{alayrac2022flamingo,moor2023medflamingo}.
LLaVA-Med further introduced biomedical visual instruction tuning by adapting a general vision-language assistant with biomedical captions and instruction-following data \cite{li2023llava_med}.
Recent state-of-the-art VLMs such as Qwen2-VL improve image understanding through dynamic resolution processing and multi-modal positional encoding, while medical VLMs such as MedGemma introduce medical-specific visual encoders to further improve the domain adaptation \cite{wang2024_qwen,medgemma}.
Despite the remarkable performance on task-specific benchmarks, these VLMs focus on a single reasoning pathway, which limits the clinical correctness.
\subsection{Multi-Agent Vision Language Models in Medical Report Generation}
Researchers have explored several strategies to combine the strengths of multiple LLMs, including ensembling (e.g., Mixture-of-Experts), knowledge distillation, and multi-agent debate (MAD) \cite{lin2026moe,lee2026distillation,tang2024medagents}.
While Mixture-of-Experts and knowledge distillation improve capacity, efficiency, or output quality, they rely on post-training or aggregation mechanisms, and thus do not explicitly support interactive, evidence-grounded reasoning at inference time.
Inspired by Minsky's Society of Mind \cite{minsky1986society}, MAD instead allows multiple agents to assume different roles, exchange intermediate analyses, and reach a final decision through discussion or consensus \cite{du2023improving,liangetal2024encouraging,chen2024reconcile}.

Compared with general-purpose multi-agent frameworks, MedAgents~\cite{tang2024medagents} extended this idea to zero-shot medical reasoning by assigning domain expert roles and using multi-round discussion before final decision making.
MDAgents~\cite{kim2024mdagents} further adapted the collaboration structure according to medical task complexity, showing that the organization of agents can affect both efficiency and accuracy.
Despite improvements over single LLMs, these models mainly rely on role-based discussion or response aggregation over already language-based evidence, which can lead to reasoning drift, low-quality feedback, and homogeneous outputs \cite{tran2025multiagent_collabortion}.
We argue that medical image understanding requires a different form of collaboration, where agents learn from specialized clinical evidence sources and their outputs are consolidated through complementary reasoning under a training-free judge mechanism.

\section{Method}
\subsection{Overview of the LLM Council}
We define SMILE as a council of LLMs specifically designed to analyze complementary attributes of liver ultrasound images. Each agent is fine-tuned on unique target datasets under a supervised fine-tuning (SFT) approach, which is illustrated in Fig. \ref{fig:method}(a). We adopt parameter-efficient low-rank adaptation (LoRA) \cite{hu2021}, where agent-specific learnable parameters are introduced while keeping the shared multi-modal backbone frozen. Once all agents have been specialized, we ask them to collaborate under the guidance of an anchor LLM, referred to as the judge, which refines the outputs of each agent and provides both criticism and a verdict for their responses. As shown in Fig. \ref{fig:method}(b), a judge can provide the final caption under a defined set of rules.

\subsection{Problem Setup}
Let $\mathcal{D} = \{(x_i, y_i)\}_{i=1}^{N}$ denote the training dataset, where $x_i$ is a liver ultrasound case and $y_i$ is the corresponding ground-truth caption. Our goal is to approximate the conditional distribution of a clinically coherent report given an ultrasound image, denoted as $p(\tilde{y} \mid x),$ where $\tilde{y}$ represents the final consensus caption generated through multi-agent reasoning and judge aggregation.

\subsection{Agent-Level Conditional Modeling}
\label{subsec:agent_training}
Let $\mathcal{D}^{(j)}=\{(x_i, y_i^{(j)})\}_{i=1}^{N}$ denote the agent-specific training dataset, 
where $y_i^{(j)}$ is the ground-truth caption 
for agent $j \in \{1,\dots,M\}$, represented as a token sequence $y_i^{(j)} = (y_{i,1}^{(j)}, \ldots,y_{i,T_i^{(j)}}^{(j)})$. Each agent $f_{\theta,\phi_j}$ employs a frozen multi-modal backbone parameterized by $\theta$ and is adapted via the LoRA module with learnable parameters $\phi_j$. By conditioning on an agent-specific prompt $\pi_j$, the conditional distribution modeled by agent $j$ is defined as:
\begin{equation}
p_{\theta,\phi_j}
\left(
y_i^{(j)} \mid x_i, \pi_j
\right)
=
\prod_{t=1}^{T_i^{(j)}}
p_{\theta,\phi_j}
\left(
y_{i,t}^{(j)} \mid x_i, \pi_j, y_{i,<t}^{(j)}
\right).
\end{equation}
At inference time, the generated caption of agent $j$ is sampled as:
\begin{equation}
\hat{y}_i^{(j)}
\sim
p_{\theta,\phi_j}
\left(
y \mid x_i, \pi_j
\right).
\end{equation}
Each agent is trained independently using the SFT approach by minimizing the token-level negative log-likelihood. Specifically, the objective for agent $j$ can be written as:
\begin{equation}
\mathcal{L}_{\mathrm{SFT}}^{(j)}
=
- \sum_{i=1}^{N}
\sum_{t=1}^{T_i^{(j)}}
\log
p_{\theta,\phi_j}
\left(
y_{i,t}^{(j)} 
\mid 
x_i, \pi_j, y_{i,<t}^{(j)}
\right).
\end{equation}

\subsection{Inference and Consolidation with a Judge}
\label{subsec:judge_methodology}
Once all agents are trained, their generated captions $\{\hat{y}_i^{(j)}\}_{j=1}^{M}$ are aggregated and criticized by a judge model. The judge is a frozen LLM parameterized by $\theta$ and conditioned on a dedicated prompt $\pi^{*}$.  We refer to $\pi^{*}$ as the consensus filtering prompt, as it instructs the judge to extract a consensus from the complementary agentic responses by filtering the non-visual grounding attributes. 
The complete consensus-filtering prompt is provided in Supp.~Note~D.

Finally, the judge models the conditional distribution as:
\begin{equation}
p_{\theta}
\left(
\tilde{y}_i 
\mid 
x_i, 
\{\hat{y}_i^{(j)}\}_{j=1}^{M}, 
\pi^{*}
\right),
\end{equation}
and produces the final consensus caption
\begin{equation}
\tilde{y}_i
\sim
p_{\theta}
\left(
y 
\mid 
x_i, 
\{\hat{y}_i^{(j)}\}_{j=1}^{M}, 
\pi^{*}
\right).
\end{equation}

\section{Experiments}
\label{sec:experiments}

\begin{table*}[!ht]
\centering
\begin{tabular}{llccccccccccc}
\toprule
\multirow{2}{*}{\textbf{Dataset}} & \multirow{2}{*}{\textbf{Model}}
& \multicolumn{4}{c}{\textbf{NLG}}
& \multicolumn{4}{c}{\textbf{CE}} 
& \multicolumn{3}{c}{\textbf{Cost}}

\\
\cmidrule(lr){3-6} \cmidrule(lr){7-10} \cmidrule(lr){11-13}
& & \textbf{R-1} & \textbf{R-L} & \textbf{M} & \textbf{BS} & \textbf{Acc} & \textbf{Prec} & \textbf{Rec} & \textbf{F1} & \textbf{P} & \textbf{L} & \textbf{FLOPs}  \\
\midrule

\multirow{5}{*}{D-1}
& Qwen-2B \cite{wang2024_qwen}      & 0.10 & 0.09 & 0.02 & 0.81 & 0.88 & 0.88 & 0.05 & 0.01 & 4.36 & 13.2 & 3.9 \\
& MedMO-4B \cite{medmo}     & 0.69 & 0.59 & 0.47 & 0.93 & 0.94  & 0.84 & 0.59 & 0.69 & 80.54 & 23.8 & 7.0 \\
& MedGem-4B \cite{medgemma}  & 0.59 & 0.48 & 0.49 & 0.90 & 0.94  & 0.78 & 0.66 & 0.72 & 77.00 & 25.6 & 7.6 \\

& MedAgents \cite{tang2024medagents}  & 0.60 & 0.45 & 0.50 & 0.90 & 0.94 &  0.77 & 0.70 & 0.73 & 55.63 & 228.9 & 81.9 \\

& \textbf{SMILE (Ours)}& \textbf{0.72} & \textbf{0.61} & \textbf{0.56} & \textbf{0.93} 
& \textbf{0.96} & 0.85 & \textbf{0.80} & \textbf{0.82} & 55.63 & 74.3 & 34.0 \\
\midrule

\multirow{5}{*}{D-2}
& Qwen-2B \cite{wang2024_qwen}      & 0.19 & 0.19 & 0.08 & 0.74 & -- & -- & -- & -- & 4.36 & 15.5 & 6.1 \\
& MedMO-4B \cite{medmo}    & 0.40 & 0.38 & 0.29 & 0.80 & -- & -- & -- & -- & 80.54 & 28.1 & 11.0 \\
& MedGem-4B \cite{medgemma}  & 0.42 & 0.39 & \textbf{0.35} & 0.78 & -- & -- & -- & -- & 77.00 & 30.2 & 11.9 \\


& MedAgents \cite{tang2024medagents}  & 0.37 & 0.33 & 0.31 & 0.77 & -- & -- & -- & -- & 55.63 & 266.1 & 116.7 \\

& \textbf{SMILE (Ours)}& \textbf{0.50} & \textbf{0.50} & 0.32 & \textbf{0.81} 
& -- & -- & -- & -- & 55.63 & 81.8 & 35.3\\
\bottomrule
\end{tabular}
\caption{
Quantitative comparison of the proposed SMILE with finetuned state-of-the-art models. We report the NLG, CE and computational cost on the test subsets of Dataset-1 and Dataset-2. The best results are highlighted in bold.
P, L, and FLOPs denote trainable parameters, inference latency, and floating-point operations, reported in millions ($10^{6}$), seconds, and trillion ($10^{12}$) operations, respectively.
}
\label{tab:ft_main_results}
\end{table*}

\subsection{Datasets}
\label{subsec:datasets}
\paragraph{Dataset-1.} 
We utilize a publicly available Liver Ultrasound dataset consisting of 1,183 liver ultrasound images and their corresponding radiology reports \cite{li2024}.
For each patient, the dataset provides two distinct liver views along with a textual description of the overall liver appearance.
We divide the dataset at the patient level into training, validation, and test sets with a 0.70/0.10/0.20 ratio. 
\paragraph{Dataset-2.} 
We retrospectively collected an independent multi-modal liver fibrosis ultrasound dataset under a protocol approved by the Institutional Review Board of the College of Medicine, The Catholic University of Korea (KC25ZISI0058).
Dataset includes 1,417 liver ultrasound images obtained from 286 patients (mean age: $55.18 \pm 16.04$ years; 105 males (36.7\%) and 181 females (63.3\%)), along with 286 corresponding clinical descriptions.
The ultrasound examinations consist of standardized liver views, including transverse and longitudinal scans of the left hepatic lobe, subcostal and intercostal scans of the right hepatic lobe, and hepatorenal comparative views.
The clinical descriptions comprise imaging findings, histopathological examination results, and fibrosis stages based on the METAVIR grading system.
The distribution of fibrosis stages ($n=286$) is as follows: F0, 34 (11.9\%); F1, 107 (37.4\%); F2, 38 (13.3\%); F3, 67 (23.4\%); F4, 40 (14.0\%).
We divide the patients into training and testing subsets with a ratio of 0.80/0.20. Additionally, we further split the training dataset, holding out 10\% of it for validation.
\subsection{Benchmarks}
\subsubsection{Baselines}
We compare SMILE with three open source vision language models, including Qwen-2B~\cite{wang2024_qwen}, MedMO-4B~\cite{medmo}, and MedGemma-4B~\cite{medgemma}.
All baselines are fine-tuned on the training subset of each benchmark dataset, following the same preprocessing and evaluation protocol to ensure a fair comparison with SMILE.

We further compare with MedAgents~\cite{tang2024medagents}, a training-free multi-agent LLM for collaborative medical reasoning.
To ensure a fair comparison, we follow the role-playing architecture and multi-stage collaboration protocol from the official implementation, including expert gathering, individual expert analysis, synthesized report generation, collaborative consultation, and final decision-making. 
Since the original MedAgents framework is designed for text-based medical reasoning, we adapt its prompts and agent roles to the liver ultrasound setting while preserving its original collaboration structure.
Specifically, the individual experts, radiology- and pathology-specialized agents, have been fine‑tuned on their respective domain-specific datasets.
The task-adapted MedAgents configuration and collaboration protocol are detailed in Supp.~Note~A.1.
\subsubsection{Evaluation Metrics}

For natural language generation (NLG), we compare the generated text with ground truth reports and calculate ROUGE-1 (R-1), ROUGE-L (R-L) \cite{lin2004rouge}, METEOR (M) \cite{banerjee2005meteor}, and BERTScore (BS) \cite{zhang2019bertscore} for assessing the linguistic quality of generated responses. 
For clinical efficacy (CE), we extract the appearance-based common keywords from generated reports and report accuracy, precision, recall, and F1-score.

We evaluate clinical correctness using closed-set visual question answering (VQA).
We derive a set of questions from appearance-based ultrasound findings and clinical attributes described in the ground truth reports of both datasets.
We report accuracy and F1-score for VQA evaluation. 

We utilize MedCLIP \cite{wang2022medclip} to encode the ultrasound images and reports into visual and textual embeddings, respectively. 
We measure the image text alignment by computing the cosine similarity between the visual embedding of each liver ultrasound image and the text embedding of its generated report. A detailed discussion on evaluation metrics is provided in Supp.~Note~B.


\subsubsection{Implementation Details}
\paragraph{Role of Agents.}
We define three agents, represented by $A$, $B$, and $C$, to decompose liver ultrasound understanding into complementary reasoning tasks.
\emph{Agent A} generates structured descriptions of appearance-based features, including echogenicity patterns, parenchymal texture, organ boundaries, and anatomical visibility. 
Its role is restricted to visually verifiable findings, without inferring pathological conclusions beyond the image evidence. 
\emph{Agent B} performs pathology-oriented diagnostic reasoning and provides higher-level interpretations related to diffuse liver disease and fibrosis-relevant findings.
\emph{Agent C} follows a VQA setting, where it answers a predefined attribute question bank with a closed answer set.
\paragraph{Training Strategy.}
We select our agents and the judge model from a pool of open-source VLMs, including Qwen-2B, MedMO-4B, and MedGemma-4B.
Specifically, we design the training procedure to utilize target captions from Dataset-1 (n = 828) and Dataset-2 (n = 228) to train \emph{Agent A} and \emph{Agent B}, respectively.
To enforce visual grounding and reduce hallucinations and response repetition, we train \emph{Agent C} on our VQA-driven questions derived from both datasets. All agents are trained independently on their respective datasets using SFT with parameter-efficient fine-tuning method (PeFT) \cite{wang2025_peft}.
\paragraph{Post-training and Hyperparameter Selection.}
During training, we select hyperparameters on the validation subsets by performing a bounded grid search over LoRA configurations. We search over rank $r \in \{4,8,16,32\}$, scale $\alpha \in \{4,8,16,32,64\}$, learning rate in $\{2\times10^{-5}, 2\times10^{-4}, 10^{-3}, 10^{-2}\}$, and batch size in $\{2,4,8\}$.
The selected configurations are MedGemma-4B for \emph{Agent A} ($r=16,\alpha=32$) and Qwen-2B for both \emph{Agents B\&C} ($r=8,\alpha=16$), all trained with learning rate $2\times10^{-4}$. LoRA is applied to all linear projections for A/B and only query/key projections for C.
A zero-shot MedGemma-4B acts as a judge during inference, and aggregates all responses as detailed in \ref{subsec:judge_methodology}.
All models are implemented in PyTorch with CUDA using publicly available model weights and official implementations. All experiments are conducted on two A6000 GPU nodes.
Additional implementation details are provided in Supp.~Notes~A and E.

\begin{table}[!t]
\centering
\begin{tabular}{llcc}
\toprule
\textbf{Dataset} & \textbf{Model} & \textbf{Accuracy} & \textbf{F1-score}
\\
\midrule

\multirow{5}{*}{D-1}
& Qwen-2B \cite{wang2024_qwen}        & 0.62 & 0.77 \\
& MedMO-4B \cite{medmo}     & 0.67 & 0.76 \\
& MedGem-4B \cite{medgemma}     & 0.53 & 0.62 \\
& MedAgents \cite{tang2024medagents}     & 0.50 & 0.41 \\
& \textbf{SMILE (Ours)} & \textbf{0.77} & \textbf{0.82} \\
\midrule

\multirow{5}{*}{D-2}
& Qwen-2B \cite{wang2024_qwen}       & 0.32 & 0.44 \\
& MedMO-4B \cite{medmo}       & \textbf{0.56} & 0.57 \\
& MedGem-4B \cite{medgemma}     & 0.46 & 0.61 \\
& MedAgents \cite{tang2024medagents}     & 0.40 & 0.39 \\
& \textbf{SMILE (Ours)} & 0.48 & \textbf{0.65} \\
\bottomrule
\end{tabular}
\caption{
VQA-based clinical verification on Dataset-1 (D-1) and Dataset-2 (D-2)
}
\label{tab:vqa_results}
\end{table}

\begin{table}[!ht]
\centering
\begin{tabular}{llcc}
\toprule
\textbf{Dataset} & \textbf{Model} & \textbf{Mean} & \textbf{Std} 
\\
\multirow{5}{*}{D-1}
& Qwen-2B \cite{wang2024_qwen}       & 0.48 & 0.014 \\
& MedMO-4B \cite{medmo}      & 0.45 & 0.009 \\
& MedGem-4B \cite{medgemma}      & 0.49 & 0.020 \\
& MedAgents \cite{tang2024medagents}      & 0.45 & 0.007 \\
& \textbf{SMILE (Ours)} & \textbf{0.51} & 0.014 \\
\midrule

\multirow{5}{*}{D-2}
& Qwen-2B \cite{wang2024_qwen}       & 0.33 & 0.010\\
& MedMO-4B \cite{medmo}       & 0.32 & 0.064 \\
& MedGem-4B \cite{medgemma}      & 0.30 & 0.009 \\
& MedAgents \cite{tang2024medagents}  & 0.32  & 0.017 \\
& \textbf{SMILE (Ours)} & \textbf{0.36} & 0.010 \\
\bottomrule
\end{tabular}

\caption{
Image text alignment between liver ultrasound images and generated reports. We report the mean and standard deviation of cosine similarity between visual and textual embeddings.
}
\label{tab:cosine_similarity_results}
\end{table}

\begin{table*}[!ht]
\centering
{
\begin{tabular}{lcccccccc}
\toprule
\textbf{Setting} 
& \textbf{R-1}
& \textbf{R-L} 
& \textbf{M} 
& \textbf{CE-F1}
& $\mathbf{P}$ \textbf{(M)}
& \textbf{Calls}
& $\mathbf{L}$ \textbf{(s)}
& \textbf{FLOPs} \\
\midrule

\multicolumn{9}{l}{\textit{Single Agents}} \\
\emph{Agent A} (Radiology Specialist) 
& 0.44\sig & 0.37\sig & 0.24\sig & 0.53 
& 38.50 & 1 & 25.6 & 7.6 \\

\emph{Agent B} (Pathology Specialist) 
& 0.25\sig & 0.24\sig & 0.05\sig & 0.03 
& 14.61 & 1 & 13.2 & 3.9 \\

\emph{Agent C} (VQA Specialist) 
& 0.18\sig & 0.14\sig & 0.06\sig & 0.26 
& 2.52 & 1 & 49.4 & 14.6 \\

\midrule
\multicolumn{9}{l}{\textit{Pairwise Agents}} \\

A + B 
& 0.70 & \textbf{0.62} & 0.50\sig & 0.70 
& 53.11 & 3 & 70.2 & 24.4 \\

A + C 
& 0.42\sig & 0.33\sig & 0.16\sig & 0.73 
& 41.02 & 3 & 58.7 & 27.9 \\

B + C 
& 0.29\sig & 0.21\sig & 0.13\sig & 0.62 
& 17.13 & 3 & 46.3 & 24.2 \\

\midrule
\multicolumn{9}{l}{\textit{Judge Input}} \\

SMILE (Judge with image only)
& 0.17\sig & 0.14\sig & 0.06\sig & 0.20
& 0.00 & 1 & 26.7 & 8.6 \\

SMILE (Judge with agent outputs only)
& 0.48\sig & 0.38\sig & 0.32\sig & 0.74
& 55.63 & 4 & 69.6 & 29.6 \\

\midrule
\multicolumn{9}{l}{\textit{Fusion Controls}} \\

SMILE (Judge with an ensembling prompt)
& 0.41\sig & 0.32\sig & 0.20\sig & 0.53
& 55.63 & 4 & 74.3 & 34.1 \\

SMILE (Judge with a Cross Attention Module)
& 0.23\sig & 0.17\sig & 0.09\sig & 0.42
& 120.81 & 4 & 68.5 & 28.6 \\

\midrule
\multicolumn{9}{l}{\textit{Council and Judge Controls}} \\
SMILE (Homogeneous Council)
& 0.68\sig & 0.55\sig & \textbf{0.57} & 0.77
& 60.97 & 4 & 84.8 & 33.1 \\
SMILE (General Agents)
& 0.30\sig & 0.21\sig & 0.13\sig & 0.72
& 0.00 & 4 & 74.3 & 34.0 \\
SMILE (Qwen-2B as a Judge) & 0.57\sig &  0.45\sig & 0.36 & 0.77
& 55.63 & 4& 59.0 & 27.7 \\
\textbf{SMILE (Ours)}
& \textbf{0.72} & 0.61 & 0.56 & \textbf{0.82}
& 55.63 & 4 & 74.3 & 34.1 \\

\bottomrule
\end{tabular}
}
\caption{
Ablation analysis of SMILE on Dataset-1. We report single agents, pairwise agents, judge-input controls, fusion controls, and judge-backbone ablations.
$P$ denotes the number of trainable parameters in millions, Calls denotes the number of model invocations per patient, $L$ denotes average inference latency per patient in seconds, and FLOPs denote floating-point operations, reported in trillions ($10^{12}$).
CE-F1 represents the clinical efficacy F1-score.
For the NLG metrics, \sig{} indicates that the complete SMILE model significantly outperforms the corresponding ablation according to a two-sided Wilcoxon signed-rank test with Holm correction ($p_{\mathrm{adj}}<0.05$). Unmarked NLG values indicate that the difference is not statistically significant. The statistical testing protocol is detailed in Supp.~Note~F.
}
\label{tab:ablation_smile_detailed}
\end{table*}

\subsection{Results}
\label{subsec:results}
\subsubsection{Medical Image Understanding}
Table~\ref{tab:ft_main_results} reports the performance of SMILE against fine‑tuned VLMs and a multi‑agent VLM baseline.
SMILE achieves the best overall performance among fine‑tuned open‑source VLMs on both datasets.
On Dataset-1, our method outperforms the strongest fine-tuned baseline, MedMO-4B, improving ROUGE-1 by 4.3\% from 0.69 to 0.72, ROUGE-L by 3.4\% from 0.59 to 0.61, and METEOR by 19.1\% from 0.47 to 0.56.
Under CE evaluation, SMILE improves the F1-score by 18.8\% from 0.69 to 0.82 compared with MedMO-4B.
Similarly, on Dataset-2, SMILE outperforms the strongest fine-tuned baseline, MedGemma-4B, improving both ROUGE-1 and ROUGE-L by 19.0\% and 28.2\%, respectively.

We further compare SMILE with MedAgents to evaluate the effectiveness of our designed framework for liver ultrasound understanding. 
On Dataset-1, SMILE improves over MedAgents by 0.12 in ROUGE-1, 0.16 in ROUGE-L, 0.06 in METEOR, and 0.09 in CE-F1.
On Dataset-2, SMILE also improves over MedAgents by 0.13 in ROUGE-1, 0.17 in ROUGE-L, and 0.01 in METEOR.

SMILE achieves stronger medical image understanding across NLG and CE metrics despite requiring fewer trainable parameters than the single VLM baselines, such as MedMO-4B and MedGemma-4B. Compared with MedAgents under the same trainable parameter count of 55.63M, SMILE requires much less inference compute, with \st{$2.7\times$} $2.4\times$ fewer
FLOPs and $3.1\times$ lower latency on Dataset-1 ($3.31\times$ and $3.25\times$ on
Dataset-2). 
These results show that the role specialization and consensus filtering improve the use of both model capacity and efficiency of inference cost (see more analysis in Supp.~Note~E).

\subsubsection{Evaluation on Visual Question Answering} Table~\ref{tab:vqa_results} summarizes closed-ended VQA performance across 837 and 108 evaluated questions for Dataset-1 and Dataset-2, respectively.
On Dataset-1, SMILE achieves the highest accuracy and F1-score, with scores of 0.77 and 0.82, respectively.
Compared with the strongest baseline by F1-score, Qwen-2B, SMILE improves F1-score from 0.77 to 0.82, corresponding to a relative improvement of 6.5\%.
It also improves over MedMO-4B from 0.76 to 0.82 and over MedGemma-4B from 0.62 to 0.82.
Similarly, on Dataset-2, SMILE achieves the highest F1-score of 0.65, improving over MedMO-4B from 0.57 to 0.65, MedGemma-4B from 0.61 to 0.65, and Qwen-2B from 0.44 to 0.65.
\subsubsection{Image Text Alignment}
Table~\ref{tab:cosine_similarity_results} reports the cosine similarity between visual embeddings from liver ultrasound images and text embeddings from generated reports.
SMILE achieves the highest image text alignment on both datasets, with a mean similarity of 0.51 on Dataset-1 and 0.36 on Dataset-2.
On Dataset-1, SMILE improves the MedGemma-4B, from 0.49 to 0.51. It also improves over MedAgents from 0.45 to 0.51.
These results suggest that the reports generated by SMILE are more closely aligned with the corresponding ultrasound images in the shared embedding space. 
\subsubsection{Ablation Studies}
\label{subsubsec:ablation}
We perform ablation studies on agent specialization, agent collaboration, judge inputs, fusion strategy, and council composition, as shown in Table~\ref{tab:ablation_smile_detailed} (see implementation details in Supp.~Note~A).
We note that each agent specializes in specific aspects of liver ultrasound knowledge; however, on its own, it struggles to achieve optimal understanding performance.
For example, \emph{Agent A} outperforms other single-agent VLMs by achieving ROUGE-L of 0.37, METEOR of 0.24, and CE-F1 of 0.53.
While \emph{Agent B} struggles to perform on-par with \emph{Agent A} because \emph{Agent A} has been finetuned on the specialized knowledge for appearance-based interpretation as compared to \emph{Agent B}, which focuses on pathology-motivated cues.
\emph{Agent C} also performs poorly as a report generator, with ROUGE-L of 0.14 and METEOR of 0.06, but its CE-F1 of 0.26 shows that the VQA agent still contributes some finding-level evidence despite weak free text generation.
\paragraph{Pairwise Agents.}
We perform pairwise ablations of our council, which shows that agent collaboration improves image understanding of liver diagnosis.
Specifically, when combined \emph{Agents A} and \emph{B}, it gives the strongest pairwise report generation scores, increasing ROUGE-L from 0.37 to 0.62 and METEOR from 0.24 to 0.50 compared with \emph{Agent A} alone.
However, its CE-F1 remains 0.70, which is further improved by adding the VQA agent, and finally enforces the constraints upon council toward clinical entity preservation.
We also observe that the A+C setting increases CE-F1 from 0.53 to 0.73 compared with \emph{Agent A} alone, even though its ROUGE-L decreases from 0.37 to 0.33.
\paragraph{Judge Ablations}

We isolate the contribution of the judge through input, fusion, and council controls: 1) using the image alone yields only 0.17 ROUGE-1 and 0.20 CE-F1, whereas providing agent outputs without visual input improves these scores to 0.48 and 0.74, showing that agent-derived responses provide stronger clinical evidence to the judge; 2) simple aggregation remains insufficient, as an ensembling prompt achieves CE-F1 of 0.53, identical to \emph{Agent A} alone, while Cross Attention fusion reaches only 0.23 ROUGE-1 and 0.42 CE-F1 despite requiring 120.81M trainable parameters; 3) a homogeneous council reaches 0.68 ROUGE-1 and 0.77 CE-F1, remaining below SMILE at 0.72 and 0.82 while requiring more trainable parameters (60.97M vs.\ 55.63M) and higher latency (84.8s vs.\ 74.3s); 4) general zero-shot agents achieve only 0.30 ROUGE-1 despite using the same four model calls, showing that additional inference alone does not guarantee better performance; and 5) replacing the MedGem-4B judge with Qwen-2B shows that specialized agents provide sufficiently informative clinical evidence for consolidation across different judge LLM.
Overall, these results, including  statistically significant improvements in NLG, suggest that SMILE benefits from complementary agent specialization and criticism-based consensus filtering rather than increased model capacity or inference cost.

\begin{figure*}[!ht]
\centering
\includegraphics[
    width=0.99\textwidth,
    height=0.45\textheight,
    keepaspectratio
]{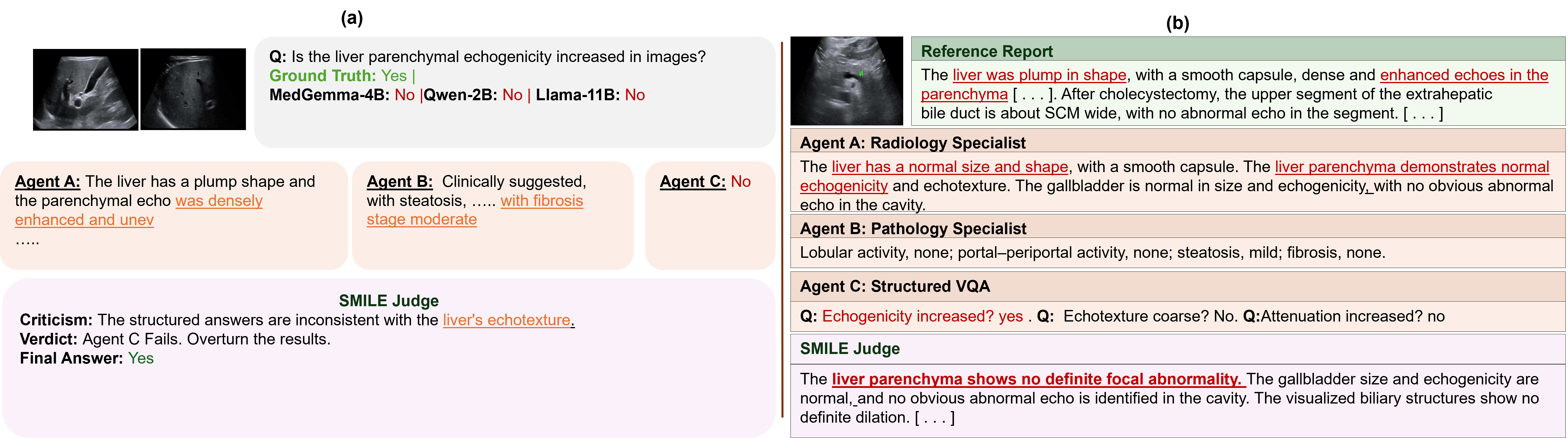}
\caption{
Qualitative analysis of SMILE. 
(a) Successful correction using complementary agent evidence. 
(b) Failure case where the judge overlooks a correct agent finding. 
Red, green, and orange denote incorrect, correct, and complementary findings, respectively.
}
\label{fig:vqa_results}
\end{figure*}


\subsubsection{Qualitative Analysis:}
To further highlight the importance of a multi-agent LLM council in providing the complementary reasoning paths for liver ultrasound, we present a case study from VQA task and a case from Report Generation Task in (a) and (b) Fig~\ref{fig:vqa_results}. 
In example (a), the image shows diffused echogenicity in the parenchymal region.
Initially, \emph{Agent C} fails to answer correctly, as do other single LLM baselines.
However, \emph{Agent A}, which reasons over grounded captions learned during fine-tuning, and \emph{Agent B}, which provides fibrosis hints based on contrast patterns, together supply the contextual information needed for the council to self-correct without any external context.
Example (b) highlights an important failure mode of the council. The reference report describes a plump liver with increased parenchymal echogenicity, whereas \emph{Agent A} incorrectly characterizes the liver as normal in shape and echogenicity. In this case, \emph{Agent C} correctly identifies increased echogenicity, yet the judge fails to incorporate this evidence and does not recover the reference liver morphology and echogenicity. This case shows that consensus aggregation can fail when correct evidence from one specialist is ignored relative to conflicting agent responses (see more case-studies in Supp.~Note~C.)


\section{Discussion}
\label{sec:dicussion}
\paragraph{Complementary roles of radiology- and pathology-aligned supervision.}
SMILE leverages two complementary sources of clinical supervision instead of relying on a single source.
Radiology-aligned supervision captures visually observable findings, including echogenicity, parenchymal texture, anatomical visibility, and lesion appearance.
Pathology-aligned supervision provides disease-level interpretive context through textual staging labels (e.g., fibrosis grade, steatosis score).
Instead of forcing a single VLM to map images directly to both visual descriptions and disease stages, we assign each supervision target to a specialized agent.
Finally, a training-free judge VLM, aligns the outputs of these agents by retaining findings supported by complementary evidence.
\paragraph{Benefits of consensus filtering over role-based aggregation.}
Role-based aggregation, as used in MedAgents, improves general medical QA through iterative expert discussion. However, it is less effective for ultrasound-specific visual reasoning because it lacks an explicit mechanism to verify whether a stated finding is grounded in the image. 
In SMILE, each agent is prompted to provide a complementary source of evidence: radiology-oriented morphology, pathology-aligned disease interpretation, and structured fact verification.
Our prompting strategy ensures that the judge does not simply merge these outputs or select the majority response.
Instead, it cross-references each candidate's findings against the complementary evidence from the other agents, as shown in Fig.~\ref{fig:vqa_results}.
We observe that findings unsupported by cross-agent evidence or contradicted by other agents are filtered out, reducing hallucination-related errors in the final output.
\paragraph{Clinical adaptability under limited-data conditions.}
SMILE improves clinical adaptability by reducing the training and inference requirements that often limit medical multi-agent systems in data-scarce radiology settings.
First, SMILE does not require end-to-end fine-tuning of a large VLM. Each agent is adapted through lightweight LoRA for a specific supervision source.
Second, SMILE does not rely on multiple rounds of agent debate. Each agent produces its report once, while a zero-shot VLM acting as the judge performs consensus filtering, reducing inference complexity and limiting repeated reasoning errors compared with multi-agent systems that require iterative discussion.
Third, SMILE enables learning from unpaired radiology and pathology supervision without requiring each ultrasound case to include both a radiology report and a pathology-based description.
Instead, agents are specialized using their respective supervision sources, and their outputs are integrated at inference through judge-guided consensus filtering.
This design makes SMILE better suited for heterogeneous clinical datasets, where data are often uni-modal, institution-dependent, and collected under different reporting protocols.
\paragraph{Limitations.}
Our results also highlight the limitations of direct fine-tuning in low-data medical VLM settings.
Although fine-tuned single models improve task adaptation, they remain sensitive to dataset bias and incomplete clinical supervision.
If all agents miss the same finding or produce the same incorrect observation, the judge may not recover the correct interpretation.
Furthermore, while we avoid iterative debate, the parallel forward pass of three specialized agents increases inference latency compared to a single model.
Future work should evaluate SMILE across larger multi-center cohorts and diverse clinical tasks to assess its broader generalization.
\section{Conclusion}
We propose SMILE, a structured multi-agent framework that enhances interpretability in liver ultrasound understanding. By leveraging agents trained on diverse information sources, our approach provides complementary context to generate visually-grounded and accurate medical reports. Experimental results demonstrate that SMILE significantly outperforms conventional methods in both image understanding and clinical diagnostic performance. Future work will focus on incorporating reinforcement learning-based policy optimization for judge decisions and developing more sophisticated collaborative rules, which could drive further breakthroughs in both natural imaging and clinical applications.
\section*{Acknowledgement}
This work was supported by the National Research Foundation of Korea (NRF) grant funded by the Korea government (MSIT) (No. RS-2024-00335185). This work was also supported by the Korea Medical Device Development Fund grant funded by the Korea government (the Ministry of Science and ICT, the Ministry of Trade, Industry and Energy, the Ministry of Health \& Welfare, the Ministry of Food and Drug Safety) (Project Number: RS-2026-25543484). 
{
    \small
    \bibliographystyle{ieeenat_fullname}
    \bibliography{references}

@article{bataller2005liver,
  title={Liver fibrosis},
  author={Bataller, Ram{\'o}n and Brenner, David A and others},
  journal={J. Clin. Invest.},
  volume={115},
  number={2},
  pages={209--218},
  year={2005},
  publisher={American Society for Clinical Investigation},
  doi={10.1172/JCI24282}
}

@article{wynn2012mechanisms,
  title={Mechanisms of fibrosis: therapeutic translation for fibrotic disease},
  author={Wynn, Thomas A and Ramalingam, Thirumalai R},
  journal={Nat. Med.},
  volume={18},
  number={7},
  pages={1028--1040},
  year={2012},
  publisher={Nature Publishing Group US New York},
  doi={10.1038/nm.2807}
}

@article{2002.21.9.1023,
author = {Tchelepi, Hisham and Ralls, Philip W. and Radin, Randall and Grant, Edward},
title = {Sonography of Diffuse Liver Disease},
journal = {J. Ultrasound Med.},
volume = {21},
number = {9},
pages = {1023-1032},
doi = {10.7863/jum.2002.21.9.1023},
year = {2002}
}

@article{lee2020deep,
  title={Deep learning with ultrasonography: automated classification of liver fibrosis using a deep convolutional neural network},
  author={Lee, Jeong Hyun and Joo, Ijin and Kang, Tae Wook and Paik, Yong Han and Sinn, Dong Hyun and Ha, Sang Yun and Kim, Kyunga and Choi, Choonghwan and Lee, Gunwoo and Yi, Jonghyon and others},
  journal={Eur. Radiol.},
  volume={30},
  number={2},
  pages={1264--1273},
  year={2020},
  publisher={Springer},
  doi={10.1007/s00330-019-06407-1}
}

@article{joo2023classification,
  title={Classification of liver fibrosis from heterogeneous ultrasound image},
  author={Joo, Yunsang and Park, Hyun-Cheol and Lee, O-Joun and Yoon, Changhan and Choi, Moon Hyung and Choi, Chang},
  journal={Ieee Access},
  volume={11},
  pages={9920--9930},
  year={2023},
  publisher={IEEE},
  doi={10.1109/ACCESS.2023.3240216}
}

@article{yang2025application,
  title={Application of large language models in disease diagnosis and treatment},
  author={Yang, Xintian and Li, Tongxin and Su, Qin and Liu, Yaling and Kang, Chenxi and Lyu, Yong and Zhao, Lina and Nie, Yongzhan and Pan, Yanglin},
  journal={Chin. Med. J.},
  volume={138},
  number={02},
  pages={130--142},
  year={2025},
  publisher={Chinese Medical Journals Publishing House Co., Ltd. 42 Dongsi Xidajie~…},
  doi={10.1097/CM9.0000000000003456}
}

@article{liu2024claude,
  title={Claude 3 Opus and ChatGPT with GPT-4 in dermoscopic image analysis for melanoma diagnosis: comparative performance analysis},
  author={Liu, Xu and Duan, Chaoli and Kim, Min-kyu and Zhang, Lu and Jee, Eunjin and Maharjan, Beenu and Huang, Yuwei and Du, Dan and Jiang, Xian},
  journal={JMIR Med. Inform.},
  volume={12},
  pages={e59273},
  year={2024},
  publisher={JMIR Publications Toronto, Canada},
  doi={10.2196/59273}
}

@article{hirosawa2023chatgpt,
  title={ChatGPT-generated differential diagnosis lists for complex case--derived clinical vignettes: diagnostic accuracy evaluation},
  author={Hirosawa, Takanobu and Kawamura, Ren and Harada, Yukinori and Mizuta, Kazuya and Tokumasu, Kazuki and Kaji, Yuki and Suzuki, Tomoharu and Shimizu, Taro},
  journal={JMIR Med. Inform.},
  volume={11},
  pages={e48808},
  year={2023},
  publisher={JMIR Publications Toronto, Canada},
  doi={10.2196/48808}
}

@article{alansari2025large,
  title={Large language models hallucination: A comprehensive survey},
  author={Alansari, Aisha and Luqman, Hamzah},
  journal={arXiv preprint arXiv:2510.06265},
  year={2025},
  doi={10.48550/arXiv.2510.06265}
}

@inproceedings{zhuo2024prosa,
  title={ProSA: Assessing and understanding the prompt sensitivity of LLMs},
  author={Zhuo, Jingming and Zhang, Songyang and Fang, Xinyu and Duan, Haodong and Lin, Dahua and Chen, Kai},
  booktitle={Findings of the Association for Computational Linguistics: EMNLP 2024},
  pages={1950--1976},
  year={2024},
  doi={10.18653/v1/2024.findings-emnlp.108}
}

@article{d2019b,
  title={B-mode ultrasound for the assessment of hepatic fibrosis: A quantitative multiparametric analysis for a radiomics approach},
  author={D’Souza, Julia C and Sultan, Laith R and Hunt, Stephen J and Schultz, Susan M and Brice, Angela K and Wood, Andrew KW and Sehgal, Chandra M},
  journal={Scientific Reports},
  volume={9},
  number={1},
  pages={8708},
  year={2019},
  publisher={Nature Publishing Group UK London},
  doi={10.1038/s41598-019-45043-z}
}

@ARTICLE{li2024,
  author={Li, Jun and Su, Tongkun and Zhao, Baoliang and Lv, Faqin and Wang, Qiong and Navab, Nassir and Hu, Ying and Jiang, Zhongliang},
  journal={IEEE Trans. Med. Imaging}, 
  title={Ultrasound Report Generation With Cross-Modality Feature Alignment via Unsupervised Guidance}, 
  year={2025},
  volume={44},
  number={1},
  pages={19-30},
  doi={10.1109/TMI.2024.3424978}}

@misc{hu2021,
      title={LoRA: Low-Rank Adaptation of Large Language Models}, 
      author={Edward J. Hu and Yelong Shen and Phillip Wallis and Zeyuan Allen-Zhu and Yuanzhi Li and Shean Wang and Lu Wang and Weizhu Chen},
      year={2021},
      eprint={2106.09685},
      archivePrefix={arXiv},
      primaryClass={cs.CL},
      doi={10.48550/arXiv.2106.09685}
}

@misc{wang2024_qwen,
      title={Qwen2-VL: Enhancing Vision-Language Model's Perception of the World at Any Resolution}, 
      author={Peng Wang and Shuai Bai and Sinan Tan and Shijie Wang and Zhihao Fan and Jinze Bai and Keqin Chen and Xuejing Liu and Jialin Wang and Wenbin Ge and Yang Fan and Kai Dang and Mengfei Du and Xuancheng Ren and Rui Men and Dayiheng Liu and Chang Zhou and Jingren Zhou and Junyang Lin},
      year={2024},
      eprint={2409.12191},
      archivePrefix={arXiv},
      primaryClass={cs.CV},
      doi={10.48550/arXiv.2409.12191
}
}

@misc{medmo,
      title={MedMO: Grounding and Understanding Multimodal Large Language Model for Medical Images}, 
      author={Ankan Deria and Komal Kumar and Adinath Madhavrao Dukre and Eran Segal and Salman Khan and Imran Razzak},
      year={2026},
      eprint={2602.06965},
      archivePrefix={arXiv},
      primaryClass={cs.CV},
      doi={10.48550/arXiv.2602.06965}
}

@article{medgemma,
  title={Medgemma technical report},
  author={Sellergren, Andrew and Kazemzadeh, Sahar and Jaroensri, Tiam and Kiraly, Atilla and Traverse, Madeleine and Kohlberger, Timo and Xu, Shawn and Jamil, Fayaz and Hughes, C{\'\i}an and Lau, Charles and others},
  journal={arXiv preprint arXiv:2507.05201},
  year={2025},
  doi={10.48550/arXiv.2507.05201}
}

@article{wang2025_peft,
  title={Parameter-efficient fine-tuning in large language models: a survey of methodologies},
  author={Wang, Luping and Chen, Sheng and Jiang, Linnan and Pan, Shu and Cai, Runze and Yang, Sen and Yang, Fei},
  journal={Artif. Intell. Rev.},
  volume={58},
  number={8},
  pages={227},
  year={2025},
  publisher={Springer},
  doi={10.1007/s10462-025-11236-4}
}

@article{friedrich2009liver,
  title={Liver fibrosis in viral hepatitis: noninvasive assessment with acoustic radiation force impulse imaging versus transient elastography},
  author={Friedrich-Rust, Mireen and Wunder, Katrin and Kriener, Susanne and Sotoudeh, Fariba and Richter, Swantje and Bojunga, Joerg and Herrmann, Eva and Poynard, Thierry and Dietrich, Christoph F and Vermehren, Johannes and others},
  journal={Radiology},
  volume={252},
  number={2},
  pages={595--604},
  year={2009},
  publisher={Radiological Society of North America, Inc.},
  doi={10.1148/radiol.2523081928}
}

@inproceedings{zhang2025_multiagent,
  title={Multi-agent reasoning for cardiovascular imaging phenotype analysis},
  author={Zhang, Weitong and Qiao, Mengyun and Zang, Chengqi and Niederer, Steven and Matthews, Paul M and Bai, Wenjia and Kainz, Bernhard},
  booktitle={International Conference on Medical Image Computing and Computer-Assisted Intervention},
  pages={429--439},
  year={2025},
  organization={Springer}
}

@article{kim2025medical,
  title={Medical hallucinations in foundation models and their impact on healthcare},
  author={Kim, Yubin and Jeong, Hyewon and Chen, Shan and Li, Shuyue Stella and Park, Chanwoo and Lu, Mingyu and Alhamoud, Kumail and Mun, Jimin and Grau, Cristina and Jung, Minseok and others},
  journal={arXiv preprint arXiv:2503.05777},
  year={2025},
  doi={10.48550/arXiv.2503.05777}
}

@article{wang2024_loss_of_context_awareness,
  title={On the loss of context-awareness in general instruction fine-tuning},
  author={Wang, Yihan and Bai, Andrew and Peng, Nanyun and Hsieh, Cho-Jui},
  journal={arXiv preprint arXiv:2411.02688},
  year={2024},
  doi={10.48550/arXiv.2411.02688}
}

@inproceedings{chen-etal-2024-towards-injecting,
    title = "Towards Injecting Medical Visual Knowledge into Multimodal {LLM}s at Scale",
    author = "Chen, Junying  and
      Gui, Chi  and
      Ouyang, Ruyi  and
      Gao, Anningzhe  and
      Chen, Shunian  and
      Chen, Guiming Hardy  and
      Wang, Xidong  and
      Cai, Zhenyang  and
      Ji, Ke  and
      Wan, Xiang  and
      Wang, Benyou",
    editor = "Al-Onaizan, Yaser  and
      Bansal, Mohit  and
      Chen, Yun-Nung",
    booktitle = "Proceedings of the 2024 Conference on Empirical Methods in Natural Language Processing",
    month = nov,
    year = "2024",
    address = "Miami, Florida, USA",
    publisher = "Association for Computational Linguistics",
    doi = "10.18653/v1/2024.emnlp-main.418",
    pages = "7346--7370",
}

@inproceedings{banerjee2005meteor,
  title={METEOR: An automatic metric for MT evaluation with improved correlation with human judgments},
  author={Banerjee, Satanjeev and Lavie, Alon},
  booktitle={Proceedings of the acl workshop on intrinsic and extrinsic evaluation measures for machine translation and/or summarization},
  pages={65--72},
  year={2005}
}

@inproceedings{lin2004rouge,
  title={Rouge: A package for automatic evaluation of summaries},
  author={Lin, Chin-Yew},
  booktitle={Text summarization branches out},
  pages={74--81},
  year={2004}
}

@article{zhang2019bertscore,
  title={Bertscore: Evaluating text generation with bert},
  author={Zhang, Tianyi and Kishore, Varsha and Wu, Felix and Weinberger, Kilian Q and Artzi, Yoav},
  journal={arXiv preprint arXiv:1904.09675},
  year={2019},
  doi={10.48550/arXiv.1904.09675}
}

@inproceedings{tang2024medagents,
  title={Medagents: Large language models as collaborators for zero-shot medical reasoning},
  author={Tang, Xiangru and Zou, Anni and Zhang, Zhuosheng and Li, Ziming and Zhao, Yilun and Zhang, Xingyao and Cohan, Arman and Gerstein, Mark},
  booktitle={Findings of the Association for Computational Linguistics: ACL 2024},
  pages={599--621},
  year={2024}
}

@inproceedings{wang2022medclip,
  title={Medclip: Contrastive learning from unpaired medical images and text},
  author={Wang, Zifeng and Wu, Zhenbang and Agarwal, Dinesh and Sun, Jimeng},
  booktitle={Proceedings of the 2022 Conference on Empirical Methods in Natural Language Processing},
  pages={3876--3887},
  year={2022}
}

@inproceedings{chen2024reconcile,
  title={Reconcile: Round-table conference improves reasoning via consensus among diverse llms},
  author={Chen, Justin and Saha, Swarnadeep and Bansal, Mohit},
  booktitle={Proceedings of the 62nd Annual Meeting of the Association for Computational Linguistics (Volume 1: Long Papers)},
  pages={7066--7085},
  year={2024}
}

@article{kim2024mdagents,
  title={Mdagents: An adaptive collaboration of llms for medical decision-making},
  author={Kim, Yubin and Park, Chanwoo and Jeong, Hyewon and Chan, Yik S and Xu, Xuhai and McDuff, Daniel and Lee, Hyeonhoon and Ghassemi, Marzyeh and Breazeal, Cynthia and Park, Hae W},
  journal={Advances in Neural Information Processing Systems},
  volume={37},
  pages={79410--79452},
  year={2024}
}

@article{tran2025multiagent_collabortion,
  title={Multi-agent collaboration mechanisms: A survey of llms},
  author={Tran, Khanh-Tung and Dao, Dung and Nguyen, Minh-Duong and Pham, Quoc-Viet and O'Sullivan, Barry and Nguyen, Hoang D},
  journal={arXiv preprint arXiv:2501.06322},
  year={2025}
}

@book{minsky1986society,
  title={Society of mind},
  author={Minsky, Marvin},
  year={1986},
  publisher={Simon and Schuster}
}

@article{estornell2024multiagent_issues,
  title={Multi-llm debate: Framework, principals, and interventions},
  author={Estornell, Andrew and Liu, Yang},
  journal={Advances in Neural Information Processing Systems},
  volume={37},
  pages={28938--28964},
  year={2024}
}

@inproceedings{becker2026stay,
  title={Stay focused: Problem drift in multi-agent debate},
  author={Becker, Jonas and Kaesberg, Lars Benedikt and Stephan, Andreas and Wahle, Jan Philip and Ruas, Terry and Gipp, Bela},
  booktitle={Findings of the Association for Computational Linguistics: EACL 2026},
  pages={5068--5102},
  year={2026}
}

@inproceedings{chen2020r2gen_firsts,
  title={Generating radiology reports via memory-driven transformer},
  author={Chen, Zhihong and Song, Yan and Chang, Tsung-Hui and Wan, Xiang},
  booktitle={Proceedings of the 2020 conference on empirical methods in natural language processing (EMNLP)},
  pages={1439--1449},
  year={2020}
}

@inproceedings{li2022cross_issues_mrg,
  title={Cross-modal clinical graph transformer for ophthalmic report generation},
  author={Li, Mingjie and Cai, Wenjia and Verspoor, Karin and Pan, Shirui and Liang, Xiaodan and Chang, Xiaojun},
  booktitle={Proceedings of the IEEE/CVF Conference on Computer Vision and Pattern Recognition},
  pages={20656--20665},
  year={2022}
}

@article{alayrac2022flamingo,
  title={Flamingo: a visual language model for few-shot learning},
  author={Alayrac, Jean-Baptiste and Donahue, Jeff and Luc, Pauline and Miech, Antoine and Barr, Iain and Hasson, Yana and Lenc, Karel and Mensch, Arthur and Millican, Katherine and Reynolds, Malcolm and others},
  journal={Advances in neural information processing systems},
  volume={35},
  pages={23716--23736},
  year={2022}
}

@inproceedings{moor2023medflamingo,
  title={Med-flamingo: a multimodal medical few-shot learner},
  author={Moor, Michael and Huang, Qian and Wu, Shirley and Yasunaga, Michihiro and Dalmia, Yash and Leskovec, Jure and Zakka, Cyril and Reis, Eduardo Pontes and Rajpurkar, Pranav},
  booktitle={Machine learning for health (ML4H)},
  pages={353--367},
  year={2023},
  organization={PMLR}
}

@article{li2023llava_med,
  title={Llava-med: Training a large language-and-vision assistant for biomedicine in one day},
  author={Li, Chunyuan and Wong, Cliff and Zhang, Sheng and Usuyama, Naoto and Liu, Haotian and Yang, Jianwei and Naumann, Tristan and Poon, Hoifung and Gao, Jianfeng},
  journal={Advances in Neural Information Processing Systems},
  volume={36},
  pages={28541--28564},
  year={2023}
}

@article{lee2026distillation,
  title={Unified reinforcement and imitation learning for vision-language models},
  author={Lee, Byung-Kwan and Hachiuma, Ryo and Ro, Yong Man and Wang, Frank and Wu, Yueh-Hua},
  journal={Advances in Neural Information Processing Systems},
  volume={38},
  pages={156508--156534},
  year={2026}
}

@article{lin2026moe,
  title={Moe-llava: Mixture of experts for large vision-language models},
  author={Lin, Bin and Tang, Zhenyu and Ye, Yang and Huang, Jinfa and Zhang, Junwu and Pang, Yatian and Jin, Peng and Ning, Munan and Luo, Jiebo and Yuan, Li},
  journal={IEEE Transactions on Multimedia},
  year={2026},
  publisher={IEEE}
}

@article{du2023improving,
  title={Improving factuality and reasoning in language models through multiagent debate},
  author={Du, Yilun and Li, Shuang and Torralba, Antonio and Tenenbaum, Joshua B and Mordatch, Igor},
  journal={arXiv preprint arXiv:2305.14325},
  year={2023}
}

@inproceedings{liangetal2024encouraging,
    title = "{Encouraging Divergent Thinking in Large Language Models through Multi-Agent Debate}",
    author = "Liang, Tian  and
      He, Zhiwei  and
      Jiao, Wenxiang  and
      Wang, Xing  and
      Wang, Yan  and
      Wang, Rui  and
      Yang, Yujiu  and
      Shi, Shuming  and
      Tu, Zhaopeng",
    editor = "Al-Onaizan, Yaser  and
      Bansal, Mohit  and
      Chen, Yun-Nung",
    booktitle = "Proceedings of the 2024 Conference on Empirical Methods in Natural Language Processing",
    year = "2024",
    publisher = "Association for Computational Linguistics",
    pages = "17889--17904",
}

@article{wang2023r2gengpt,
  title={R2gengpt: Radiology report generation with frozen llms},
  author={Wang, Zhanyu and Liu, Lingqiao and Wang, Lei and Zhou, Luping},
  journal={Meta-Radiology},
  volume={1},
  number={3},
  pages={100033},
  year={2023},
  publisher={Elsevier}
}

@inproceedings{wang2022medical,
  title={A medical semantic-assisted transformer for radiographic report generation},
  author={Wang, Zhanyu and Tang, Mingkang and Wang, Lei and Li, Xiu and Zhou, Luping},
  booktitle={International Conference on Medical Image Computing and Computer-Assisted Intervention},
  pages={655--664},
  year={2022},
  organization={Springer}
}
}

\clearpage
\onecolumn
\appendix
\setlength{\textfloatsep}{7pt plus 2pt minus 2pt}
\setlength{\floatsep}{7pt plus 2pt minus 2pt}
\setlength{\intextsep}{7pt plus 2pt minus 2pt}
\renewcommand{\topfraction}{0.95}
\renewcommand{\bottomfraction}{0.90}
\renewcommand{\textfraction}{0.05}
\renewcommand{\floatpagefraction}{0.80}

\definecolor{refgreen}{RGB}{15,122,83}
\definecolor{agentorange}{RGB}{176,84,10}
\definecolor{judgepurple}{RGB}{114,60,180}
\definecolor{vqablue}{RGB}{23,93,196}
\definecolor{badred}{RGB}{179,38,30}
\definecolor{inkgrey}{RGB}{85,93,109}
\definecolor{sysgrey}{RGB}{55,65,81}

\providecommand{\SYS}{SMILE}
\providecommand{\gd}[1]{\textcolor{refgreen}{\textbf{#1}}}
\providecommand{\bd}[1]{\textcolor{badred}{\textbf{#1}}}
\providecommand{\lost}[1]{\textcolor{badred}{\uline{#1}}}
\providecommand{\elide}{{\color{inkgrey}[\,\ldots\,]}}
\providecommand{\gnote}[1]{{\color{inkgrey}\itshape #1}}
\providecommand{\slotph}[1]{\textcolor{vqablue}{$\langle$\textit{#1}$\rangle$}}
\providecommand{\okmark}{\textcolor{refgreen}{\checkmark}}
\providecommand{\badmark}{\textcolor{badred}{$\times$}}

\ifcsname rolebox\endcsname\else
\newtcolorbox{rolebox}[3][]{%
  colback=#2!4!white,colframe=#2!55!black,
  colbacktitle=#2!10!white,coltitle=#2!40!black,
  title={\sffamily\bfseries\scriptsize #3},
  boxrule=0.35pt,arc=1.4pt,
  left=4pt,right=4pt,top=1.6pt,bottom=1.8pt,
  toptitle=1.2pt,bottomtitle=1.1pt,
  before skip=1.6pt,after skip=1.6pt,#1}
\fi

\setcounter{figure}{0}
\setcounter{table}{0}
\renewcommand{\thefigure}{S\arabic{figure}}
\renewcommand{\thetable}{S\arabic{table}}
\renewcommand{\theHfigure}{S\arabic{figure}}
\renewcommand{\theHtable}{S\arabic{table}}

\providecommand{\suppnote}[2]{%
  \FloatBarrier
  \par\medskip
  \refstepcounter{section}%
  \setcounter{subsection}{0}%
  \section*{Supplementary Note \thesection: #1}%
  \label{#2}%
}

\raggedbottom

\begin{center}
  {\Large\textbf{\thetitle}\par}
  \vspace{0.5em}
  {\Large Supplementary Material\par}
\end{center}
\vspace{1.0em}

\suppnote{Baseline and Control Implementations}{note:baselines}

\subsection{Task Adapted MedAgents}
\label{sec:medagents}

MedAgents~\cite{tang2024medagents} is originally formulated as a zero-shot
collaborative medical reasoning framework. For a fair task-specific comparison,
we instantiate the three expert roles corresponding to the SMILE radiology,
pathology, and VQA specialists using the same adapted models employed in our
council. Any additional experts recruited by the MedAgents protocol retain their
original zero-shot behavior. The coordination and discussion procedure is also
preserved from the original implementation. This design provides both methods
with comparable task-specific specialist knowledge while retaining their
respective collaboration mechanisms.

\paragraph{Task adaptation.}
The three adapted experts use the same backbones and LoRA configurations as
their corresponding SMILE agents. Their combined trainable parameter budget is
$55.63$M, which matches the SMILE council. The remaining MedAgents experts and
coordination components receive no task-specific adaptation.

\begin{table}
\centering
\footnotesize
\begin{tabular}{@{}ll@{}}
\toprule
Setting & Value \\
\midrule
Adapter & LoRA \\
Radiology expert & MedGemma-4B, $r{=}16$ \\
Pathology expert & Qwen2-VL-2B, $r{=}8$ \\
VQA expert & Qwen2-VL-2B, restricted $r{=}8$ \\
Additional MedAgents experts & Zero shot \\
Coordinator / moderator & Frozen \\
Optimiser & AdamW, weight decay $0.01$ \\
Learning rate & $2\times10^{-4}$ \\
Epochs / batch & $3$ / $1$ with gradient accumulation \\
Trainable budget & $55.63$M \\
Maximum new tokens & $512$ \\
\bottomrule
\end{tabular}
\caption{Hyperparameters of Task-Adapted MedAgents}
\label{tab:medagent_settings}
\end{table}

\paragraph{Collaboration protocol.}
We retain the complete MedAgents inference procedure, which consists of expert
recruitment, independent analysis, collaborative revision, and final synthesis.
The three task-adapted specialists participate within the recruited expert panel,
while any remaining recruited experts continue to operate in the original
zero-shot setting. During collaborative revision, each participating expert
receives the current responses of the other experts before producing its updated
assessment. The final moderator then synthesizes the accumulated discussion into
the output. All model invocations required by this protocol are included in the
computational accounting reported in Table~1 of the main paper.

At an identical trainable parameter budget, the resulting MedAgents
implementation requires $2.4\times$ the FLOPs and $3.1\times$ the latency of
SMILE on Dataset-1. The comparison therefore evaluates the two collaboration
strategies under matched task-specific adaptation capacity while accounting for
their complete inference procedures.

\paragraph{Reproducibility.}
We will release the task-adapted MedAgents implementation together with the
SMILE training and inference code through the project repository with the
camera-ready version.

\subsection{Cross-Attention Fusion}
\label{sec:xattn}

This control replaces prompt-level consolidation with representation-level fusion, following
the gated cross-attention mechanism introduced in Flamingo~\cite{alayrac2022flamingo}. The
judge backbone remains the MedGemma-4B model used by \SYS{}, but it is trained here rather
than frozen.

\paragraph{Architecture.}
The three specialist outputs are encoded once by a frozen Bio\_ClinicalBERT text encoder,
and a learned agent-type embedding is added to each stream so that the judge can distinguish
its source. A Q-Former-style resampler then compresses the encoded evidence into $8$ latent
queries at an internal width of $512$. Three gated cross-attention blocks are inserted into
the MedGemma-4B decoder at layers $8$, $17$, and $25$ of $34$, each computing
$h \leftarrow h + \tanh(\alpha_l)\cdot \mathrm{CrossAttn}(q{=}h,\, kv{=}\text{evidence})$.
Every gate $\alpha_l$ is initialised to zero, so that $\tanh(\alpha_l){=}0$ at the first
optimisation step and the model is numerically identical to a plain LoRA baseline. Any
subsequent divergence therefore measures how much evidence the gates admit, and it does so
independently of any caption metric.

\paragraph{Training and budget.}
The backbone is held frozen in $4$-bit precision and adapted with LoRA $r{=}16$, trained for
one epoch on Dataset-1 with the fusion module in fp32 and the language model in bf16. The
trainable budget comprises $26.68$M for the fusion module, covering the resampler, the three
cross-attention blocks and their gates, together with $38.50$M for the LoRA adapter. The
trained judge therefore accounts for $65.18$M parameters, and the control totals $120.81$M
once the $55.63$M specialist adapters are included. Constraining the cross-attention to an
internal width of $512$ is what keeps the module comparable in size to the council, since an
unconstrained variant operating at the full decoder width of $2560$ requires $211.82$M
parameters for the fusion module alone. This control reaches $0.23$ ROUGE-1 and $0.42$
CE-F1, against $0.72$ and $0.82$ for \SYS{}, and it tests whether representation-level
fusion on its own can reproduce rule-constrained consensus consolidation.

\subsection{Council Controls}
\label{sec:council_controls}

We construct two council variants to examine specialist supervision and backbone
heterogeneity while preserving the overall SMILE inference structure.

\paragraph{General-agent council.}
The general-agent council preserves the same three agent roles, four model
invocations, role prompts, and MedGemma-4B judge used by SMILE. The specialist
agents receive no task-specific LoRA adaptation and instead operate directly
from their pretrained checkpoints under the corresponding role instructions.
This control retains the council structure and inference budget while removing
task-specific specialist supervision. It therefore evaluates the contribution
provided by pretrained zero-shot agents and role prompting alone.

\paragraph{Homogeneous council.}
The homogeneous council retains task-specific specialist supervision and the
same judge-mediated consensus procedure as SMILE, but all specialist roles are
instantiated from the MedGemma-4B backbone family. Each role is adapted using
its corresponding supervision source. The radiology, pathology, and VQA roles
therefore remain functionally specialized, while backbone diversity is removed.
This control enables the contribution of role-specific adaptation to be examined
under a common backbone family.

The general and homogeneous councils use the same final MedGemma-4B judge and
the same consensus-filtering procedure as SMILE. Their performance, trainable
parameters, model calls, latency, and FLOPs are reported in Table~4 of the main
paper.
\newpage
\suppnote{Clinical Evaluation Protocol}{note:evaluation}

\label{sec:metrics}

We use complementary automatic metrics for report similarity, clinical-entity
preservation, finding-level verification, and image--text alignment. These metrics are not
interpreted as direct measures of clinical factuality, which would require finding-level
clinical annotation.

\subsection{Natural language generation}
ROUGE-1, ROUGE-L~\cite{lin2004rouge}, METEOR~\cite{banerjee2005meteor} and
BERTScore~\cite{zhang2019bertscore} are used as \emph{report-similarity} measures: they
quantify lexical and embedding-level agreement between a generated report and its
reference. We interpret them as report-similarity measures rather than direct evidence of
clinical correctness; Table~\ref{tab:failure} illustrates this limitation.

\subsection{Clinical Efficacy}
\label{sec:ce}

Clinical efficacy (CE) measures whether a generated report preserves clinically relevant
entities from the reference report.

\paragraph{Entity vocabulary.}
We start from the $16$ appearance-based seed terms provided with Dataset-1. Several seed
terms refer to the same underlying clinical concept and are therefore consolidated into
canonical entities. To account for lexical variation in radiology reporting, we use an LLM
to propose morphological variants, plural forms, adjectival forms, and common clinical
abbreviations for each seed term. Every proposed variant is subsequently verified against
the Dataset-1 report corpus before inclusion. This process yields $13$ canonical entities
and 100 accepted surface variants:

\begin{table}[H]
\centering
\footnotesize
\begin{tabular}{@{}ll@{}}
\toprule
Canonical entity & Surface variants \\
\midrule
portal vein   & portal vein, main portal vein, portal venous \\
bile duct     & bile duct(s), intrahepatic bile ducts, biliary \\
gallbladder   & gallbladder, GB \\
pancreas      & pancreas, pancreatic \\
spleen        & spleen, splenic \\
parenchyma    & parenchyma, parenchymal, parenchymal echo, \\
              & spleen/pancreas parenchymal \\
echogenicity  & echogenicity, echogenic, echogenicities \\
echotexture   & echotexture, echotextural, coarse, \\
              & heterogeneous, heterogeneity \\
attenuation   & attenuation, acoustic attenuation, \\
              & posterior attenuation \\
capsule       & capsule, capsular \\
lesion        & lesion(s), space-occupying \\
cyst          & cyst, cysts, cystic \\
nodule        & nodule, nodules \\
\bottomrule
\end{tabular}
\caption{\textbf{Clinical entity vocabulary used for CE evaluation.}
Representative surface variants are shown for the $13$ canonical entities derived
from the Dataset-1 seed terminology. The complete vocabulary contains $100$
accepted surface variants.}
\label{tab:ce_vocabulary}
\end{table}

\paragraph{Severity handling.}
CE is evaluated at the canonical-entity level. Severity modifiers remain in the report text but are not scored as independent entities. CE therefore measures entity preservation rather than agreement in severity.

\paragraph{Matching and scoring.}
Each report is converted into a binary presence vector over the canonical entities. An entity is marked as present when any accepted surface variant occurs in the lower-cased, whitespace-normalised text. Accuracy, precision, recall, and F1-score are computed between the reference and generated vectors using micro aggregation over entities and reports. The current CE protocol is presence-based and polarity-agnostic. For example, both ``no dilation of the bile ducts'' and ``the bile ducts were dilated'' register the \texttt{bile duct} entity. CE should therefore be interpreted as preservation of clinically relevant report content and considered together with the closed-set VQA evaluation, which tests finding-level polarity.

\subsection{Closed-Set Visual Question Answering}
\label{sec:vqa}

VQA provides finding-level verification of clinically defined ultrasound attributes. For each dataset, three predefined questions are derived from the corresponding reference findings and answered independently for each patient:

\begin{table}
\centering
\footnotesize
\begin{tabular}{@{}p{0.13\linewidth}p{0.78\linewidth}@{}}
\toprule
Dataset & Question \\
\midrule
\multirow{3}{*}{D-1}
 & Is the liver parenchymal echogenicity diffusely increased in the imaged portion? \\
 & Is the hepatic echotexture coarse or heterogeneous in the imaged portion? \\
 & Is posterior acoustic attenuation increased in the imaged portion? \\
\addlinespace[2pt]
\multirow{3}{*}{D-2}
 & Is steatosis or fatty change reported for the liver, and is it visually assessable? \\
 & Is cirrhosis reported, and are cirrhotic morphological features visually assessable? \\
 & Is a focal hepatic lesion reported, and is it visually assessable? \\
\bottomrule
\end{tabular}
\caption{List of closed-ended Questions for VQA}
\label{tab:vqa_questions}

\end{table}

Answers are restricted to $\{\textrm{yes},\textrm{no},\textrm{not assessable}\}$.  Accuracy is computed over the evaluable patient-question pairs, while F1-score is obtained by macro aggregation across questions and patients. We report both metrics because the answer distributions are imbalanced across datasets and question categories. Qualitative examples are provided in Fig.~\ref{fig:vqa_results}.

\subsection{Image--Text Alignment}
\label{sec:alignment}

Cross-modal similarity provides an auxiliary measure of whether the semantic content of a generated report remains aligned with the source image. Such alignment is relevant to multimodal downstream applications, including image-report retrieval and representation learning. We therefore compute cosine similarity between the ultrasound image embedding and generated-report embedding using the fixed MedCLIP~\cite{wang2022medclip} image-text encoder.

MedCLIP was pretrained using MIMIC-CXR and CheXpert chest radiograph data rather than liver ultrasound, and we therefore expect limited sensitivity to fine-grained sonographic differences. This domain mismatch may partly explain the relatively small absolute differences observed between systems. We retain MedCLIP because it provides a publicly available medical contrastive image-text representation and has been evaluated for both zero-shot prediction and image-text retrieval~\cite{wang2022medclip}. All methods are evaluated using the same frozen encoder, so we interpret the resulting values only as a relative auxiliary alignment measure rather than as evidence of clinical correctness. ... Future work can evaluate ultrasound-specific or SigLIP-style medical contrastive
encoders once validated representations for this modality become available.

\suppnote{Success and Failure Analysis}{note:cases}

\subsection{Successful Consensus Correction}
Table~\ref{tab:success} shows representative cases in which conflicting specialist
evidence is reconciled and the judge recovers clinically relevant reference findings.
\begin{table}[H]
\centering
\scriptsize
\noindent
\begin{minipage}[t]{0.150\textwidth}
  \vspace{0pt}
  \includegraphics[width=\linewidth]{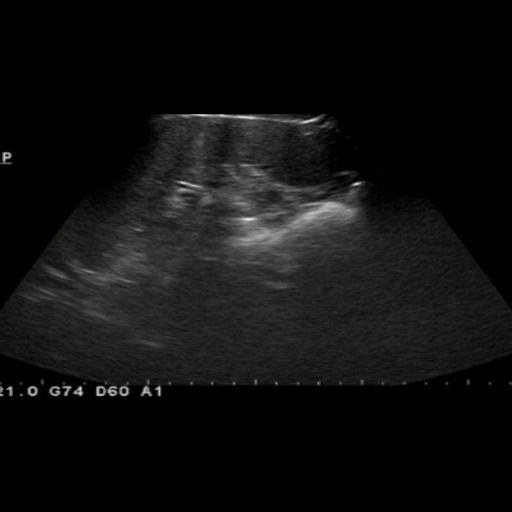}
  \par\vspace{1.5pt}
  {\centering\sffamily\bfseries Patient 225145\par}
\end{minipage}\hfill
\begin{minipage}[t]{0.835\textwidth}
\vspace{0pt}
\begin{rolebox}{refgreen}{Reference report (ground truth)}
\scriptsize The liver has \gd{a plump shape}, a smooth capsule, \gd{dense and enhanced
parenchymal echoes}, \gd{slightly attenuated backfield echoes}, unclear display of the
portal vein system, and enhanced liver and kidney echo contrast. No dilation of
intrahepatic bile ducts was found. The inner diameter of the main portal vein is within
the normal range. \elide
\end{rolebox}

\begin{rolebox}{agentorange}{Agent A : Radiology specialist response}
\scriptsize
The liver has a \bd{normal size and shape}. The capsule is
\bd{not clearly visualized}. The liver parenchyma demonstrates
\bd{normal echogenicity} with a
\bd{relatively homogeneous echotexture}. Posterior acoustic attenuation
is \bd{not evident}. \elide
\end{rolebox}
\noindent
\begin{minipage}[t]{0.505\linewidth}
\begin{rolebox}{agentorange}{Agent B : pathology specialist}
\scriptsize Lobular activity, mild; portal--periportal activity, none; steatosis, none;
fibrosis, none.
\end{rolebox}
\end{minipage}\hfill
\begin{minipage}[t]{0.475\linewidth}
\begin{rolebox}{vqablue}{Agent C : structured VQA}
\scriptsize Echogenicity increased? \gd{yes}\; $\cdot$\; Echotexture coarse? yes\;
$\cdot$\; Attenuation increased? \bd{no} \gnote{(reference: yes)}
\end{rolebox}
\end{minipage}
\begin{rolebox}{judgepurple}{\SYS{} judge}
\scriptsize
The hepatic parenchyma demonstrates
\gd{increased echogenicity} with a \gd{coarse and heterogeneous appearance}. Liver size and
capsular contour cannot be confidently characterized from the available views. Posterior
acoustic attenuation is not clearly demonstrated.
\elide
\end{rolebox}
\end{minipage}

\vspace{5pt}
\hrule height 0.2pt
\vspace{5pt}

\noindent
\begin{minipage}[t]{0.150\textwidth}
  \vspace{0pt}
  \includegraphics[width=\linewidth]{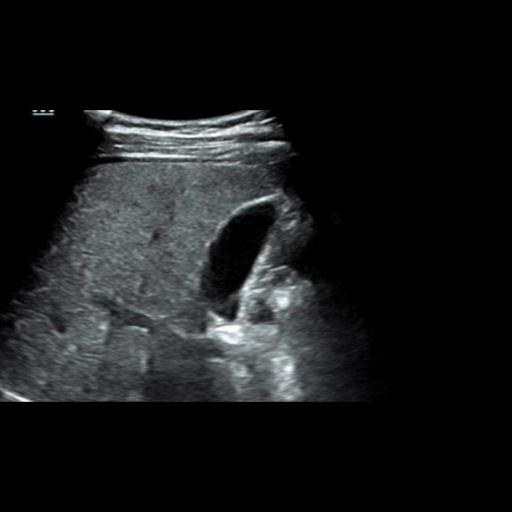}
  \par\vspace{1.5pt}
  {\centering\sffamily\bfseries Patient 219190\par}
\end{minipage}\hfill
\begin{minipage}[t]{0.835\textwidth}
\vspace{0pt}
\begin{rolebox}{refgreen}{Reference report (ground truth)}
\scriptsize The liver has \gd{a plump shape}, \gd{a smooth capsule}, \gd{dense} and uneven
parenchymal echoes, \gd{attenuated rear echoes}, enhanced liver and kidney echoes, and
unclear display of the portal venous system. \elide
\end{rolebox}
\begin{rolebox}{agentorange}{Agent A : Radiology specialist}
\scriptsize
The liver has a \bd{normal size and shape}, with the capsule \bd{not clearly visualized}.
The parenchymal echoes are \bd{relatively homogeneous}, and posterior attenuation is
\bd{unremarkable}. The portal venous system shows no obvious
abnormality. \elide
\end{rolebox}
\noindent
\begin{minipage}[t]{0.505\linewidth}
\begin{rolebox}{agentorange}{Agent B : Pathology specialist}
\scriptsize Steatohepatitis, mild; lobular activity, none; portal--periportal activity,
none; fibrosis, none; steatosis, none.
\end{rolebox}
\end{minipage}\hfill
\begin{minipage}[t]{0.475\linewidth}
\begin{rolebox}{vqablue}{Agent C : structured VQA}
\scriptsize Echogenicity increased? \gd{yes}\; $\cdot$\; Echotexture coarse? yes\;
$\cdot$\; Attenuation increased? \bd{no} \gnote{(reference: yes)}
\end{rolebox}
\end{minipage}
\begin{rolebox}{judgepurple}{\SYS{} Judge : Final Response}
\scriptsize A \gd{plump shape} of the liver is present, with a \gd{smooth capsule}.
The liver parenchyma demonstrates \gd{dense and uneven echoes}, with
\gd{attenuated rear echoes}. The liver and kidney echoes are enhanced,
and the portal venous system is not clearly displayed. \elide
\end{rolebox}
\end{minipage}

\caption{\textbf{Qualitative case studies (success): the judge overrides erroneous
specialist readings (patients 225145 and 219190, Dataset-1).} In both cases the radiology specialist reads the liver as normal and the VQA specialist answers the posterior-attenuation question incorrectly (\textcolor{badred}{\textbf{red}}), yet the consolidated report re-establishes the reference findings marked in \textcolor{refgreen}{\textbf{green}}, namely increased parenchymal echogenicity, a coarse or heterogeneous echo-texture, and the plump hepatic contour, because the remaining evidence outputs contradict the erroneous reading. \elide\ marks elided template organ sentences (bile ducts, portal vein, gallbladder, pancreas, spleen).}
\label{tab:success}
\end{table}

\FloatBarrier
\subsection{Failure Modes}
Table~\ref{tab:failure} presents the complementary failure mode: the correct finding or
clinical status is absent from the available specialist evidence and therefore remains
incorrect or missing after consolidation.

\begin{table}[H]
\centering
\scriptsize

\noindent
\begin{minipage}[t]{0.892\textwidth}
\vspace{0pt}

\begin{minipage}[t]{0.20\linewidth}
  \vspace{0pt}
  \includegraphics[width=\linewidth]{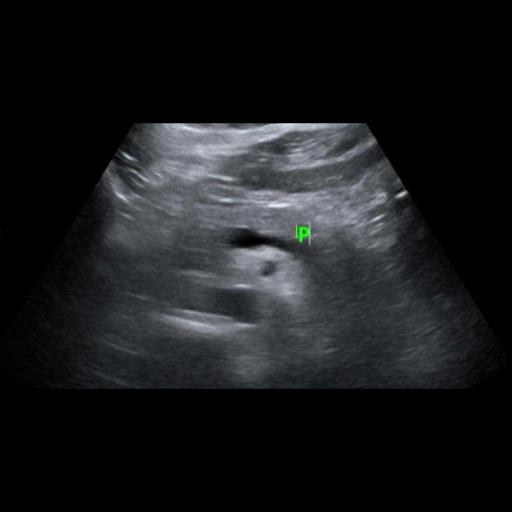}
  \par\vspace{3.5pt}
  {\centering\sffamily\bfseries Patient 216747\par}
\end{minipage}\hfill
\begin{minipage}[t]{0.78\linewidth}
\vspace{0pt}
\begin{rolebox}{refgreen}{Reference report}
\scriptsize
The liver was plump in shape, with a smooth capsule, dense and enhanced
echoes in the parenchyma \elide\
\lost{After cholecystectomy}, the upper segment of the extrahepatic
bile duct is about \texttt{\_SCM\_} wide, with no abnormal echo in the segment.
\elide
\end{rolebox}
\end{minipage}

\begin{rolebox}{agentorange}{Agent A: Radiology Specialist}
\scriptsize
The liver has a normal size and shape, with a smooth capsule.
The liver parenchyma demonstrates normal echogenicity and echotexture.
The \bd{gallbladder is normal in size and echogenicity}, with no obvious
abnormal echo in the cavity. \elide
\end{rolebox}

\noindent
\begin{minipage}[t]{0.505\linewidth}
\begin{rolebox}{agentorange}{Agent B: Pathology specialist}
\scriptsize
Lobular activity, none; portal--periportal activity, none;
steatosis, mild; fibrosis, none.
\end{rolebox}
\end{minipage}\hfill
\begin{minipage}[t]{0.475\linewidth}
\begin{rolebox}{vqablue}{Agent C: Structured VQA}
\scriptsize
Echogenicity increased? yes\; $\cdot$\; Echotexture coarse? no\;
$\cdot$\; Attenuation increased? no
\end{rolebox}
\end{minipage}

\begin{rolebox}{judgepurple}{\SYS{} Judge : Final Response}
\scriptsize
The liver parenchyma shows no definite focal abnormality.
The \bd{gallbladder size and echogenicity are normal}, and no obvious
abnormal echo is identified in the cavity. The visualized biliary
structures show no definite dilation. \elide
\end{rolebox}


\end{minipage}

\vspace{5pt}
\hrule height 0.2pt
\vspace{5pt}

\noindent
\begin{minipage}[t]{0.892\textwidth}
\vspace{0pt}

\begin{minipage}[t]{0.20\linewidth}
  \vspace{0pt}
  \includegraphics[width=\linewidth]{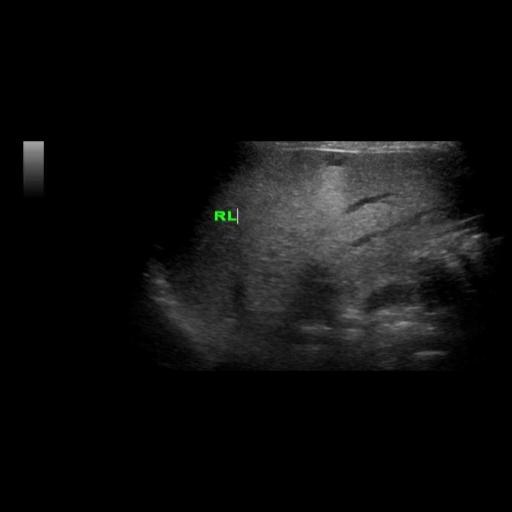}
  \par\vspace{3.5pt}
  {\centering\sffamily\bfseries Patient 235520\par}
\end{minipage}\hfill
\begin{minipage}[t]{0.78\linewidth}
\vspace{0pt}

\begin{rolebox}{refgreen}{Reference report}
\scriptsize
The liver is plump in shape and has a smooth capsule.
\lost{A cystic structure can be seen in the right lobe of the liver,
about 2DS in size, with a clear border and enhanced rear echo.}
The echo of the remaining parenchyma is \lost{finely enhanced}. \elide
\end{rolebox}

\end{minipage}

\begin{rolebox}{agentorange}{Agent A: Radiology Specialist}
\scriptsize
The liver has a normal size and shape, with a smooth capsule.
The liver parenchyma demonstrates relatively homogeneous echotexture.
There is \bd{no evidence of a focal lesion} in the visualized liver
parenchyma. \elide
\end{rolebox}

\noindent
\begin{minipage}[t]{0.505\linewidth}
\begin{rolebox}{agentorange}{Agent B: Pathology specialist}
\scriptsize
Lobular activity, none; portal--periportal activity, mild;
steatosis, none; fibrosis, none.
\end{rolebox}
\end{minipage}\hfill
\begin{minipage}[t]{0.475\linewidth}
\begin{rolebox}{vqablue}{Agent C: Structured VQA}
\scriptsize
Echogenicity increased? yes\; $\cdot$\; Echotexture coarse? no\;
$\cdot$\; Attenuation increased? no
\end{rolebox}
\end{minipage}

\begin{rolebox}{judgepurple}{\SYS{} Judge : Final Response}
\scriptsize
The liver parenchyma appears relatively homogeneous, with
\bd{no focal lesion identified} in the available views.
No additional focal abnormality is evident in the visualized hepatic
parenchyma. \elide
\end{rolebox}

{\scriptsize\color{inkgrey}\sffamily
\textbf{Failure locus:} the right-lobe cystic structure is missed by the specialist and
remains absent from the judge output. The structured VQA questions assess diffuse
parenchymal attributes and do not interrogate focal lesions.}

\end{minipage}

\vspace{2pt}

\caption{\textbf{Qualitative case studies (failure): correct findings absent from the
available evidence propagate through consensus (patients 216747 and 235520, Dataset-1).} A prior
cholecystectomy (top) and a right-lobe cystic structure (bottom) are recorded in the
reference (\textcolor{badred}{\uline{underlined}}) but the correct findings are unavailable in the specialist evidence, so the
consolidated report contradicts or omits them (\textcolor{badred}{\textbf{red}}). This
delimits the mechanism: consensus filtering is effective when corroborating evidence is available but remains
limited when the correct attribute is absent from all specialist outputs. Because the surrounding
template text is reproduced correctly, $n$-gram overlap with the reference is largely
unaffected, illustrating the limitation of report-similarity metrics
(Sec.~\ref{sec:metrics}). \texttt{\_SCM\_} is the corpus' masked-measurement token.}

\label{tab:failure}
\end{table}

\FloatBarrier
\subsection{Qualitative VQA Examples}
Figure~\ref{fig:vqa_results} provides representative closed-set VQA predictions from both
datasets.

\begin{figure}[H]
\centering
\includegraphics[width=0.65\textwidth]{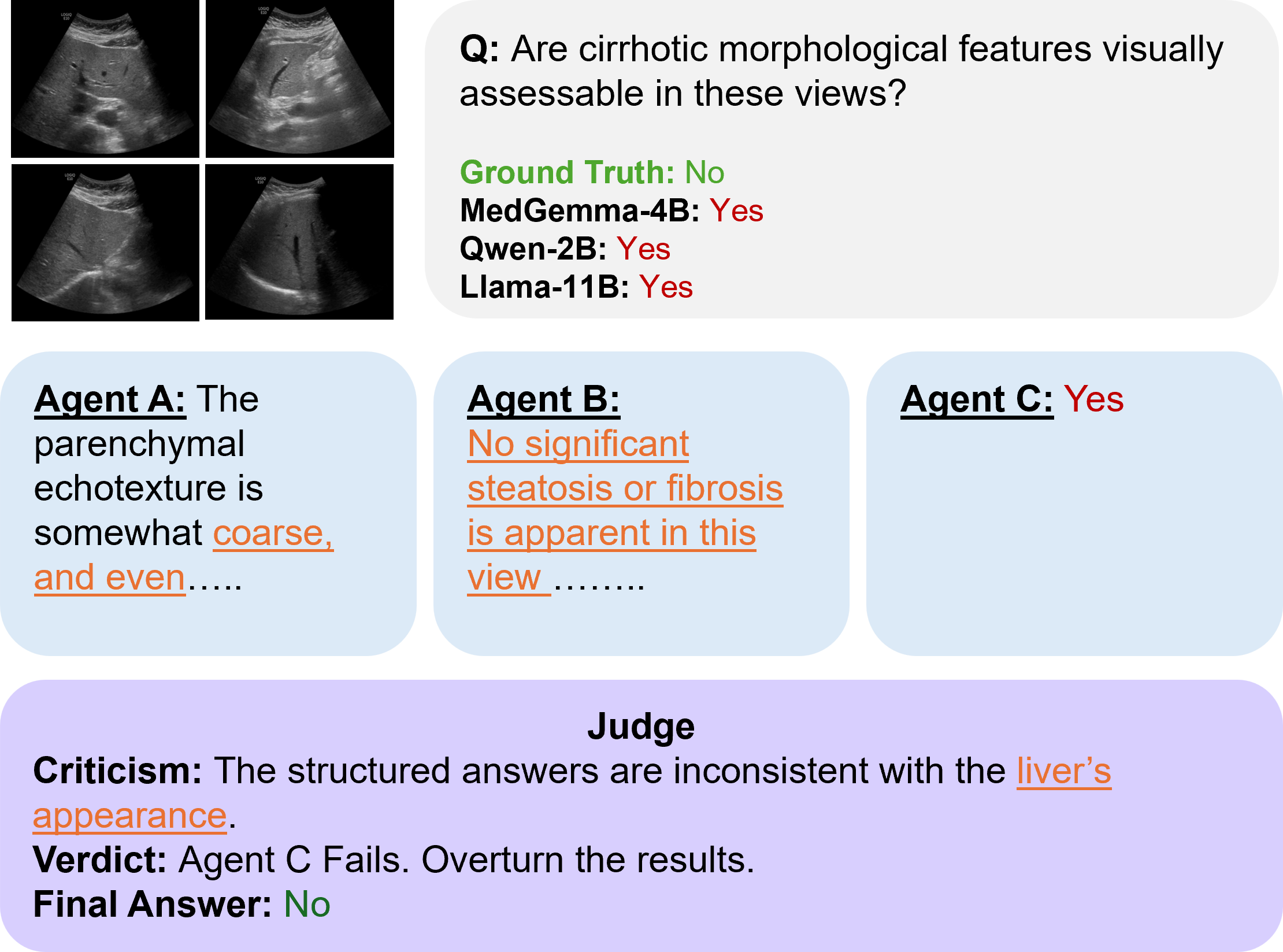}
\caption{\textbf{Qualitative case studies for closed-set VQA.} For each image and question
we compare the responses generated by our multi-agent council against the reference
answer. Red indicates incorrect predictions, green indicates correct predictions, and
orange highlights complementary information contributed by individual agents during the
reasoning process. Examples (a) and (b) are taken from Dataset-1 and Dataset-2,
respectively.}
\label{fig:vqa_results}
\end{figure}
\FloatBarrier

\suppnote{Consensus-Filtering Judge Prompt}{note:prompt}

\label{sec:prompt}

Figure~\ref{fig:prompt} reproduces the complete criticism-based consensus-filtering
instruction given to the \SYS{} judge, exactly as issued at inference time: the judge
receives both views, the candidate reports of Agents A and B, and Agent C's structured
answers; it must audit itself against three rule families (R1--R3), criticise each source
separately, and return the final caption as strict JSON. The prompt makes the consolidation rules explicit: the mandatory incorporation rule
projects Agent C's three parenchymal attributes into the final caption, constraint~4 strips
Candidate B's non-imaging pathology content, and the per-source criticisms force the judge
to commit to a verdict before writing the consolidated report.

\begin{figure}[H]
\centering
\scriptsize
\begin{rolebox}{sysgrey}{System}
\scriptsize\ttfamily
You are a senior abdominal radiologist acting as a judge. You consolidate candidate
descriptions into one final liver ultrasound caption. You must follow the constraints
strictly and output valid JSON only.
\end{rolebox}

\begin{rolebox}{vqablue}{User --- consensus-filtering instruction (verbatim; per-case content in
\textcolor{vqablue}{$\langle$slots$\rangle$})}
\scriptsize\ttfamily
\textbf{Context}\\
You are given TWO ultrasound views for the same patient in this order: view1 then view2.
Candidate A was generated from these images. Candidate B was generated from these images.\\[2.5pt]
\textbf{Hard constraints for the final caption}\\
1.~Do not mention biopsy, histopathology, pathology, laboratory tests, etiology,
symptoms, fibrosis stage, or grading systems.\\
2.~Do not claim a diagnosis.\\
3.~Do not mention agents, prompts, structured answers, VQA, or the words yes, no,
not assessable.\\
4.~Candidate B may contain non imaging content. You must ignore any non imaging content
from Candidate B.\\
5.~Only describe a structure as normal if it is clearly visible in the provided views.\\
6.~Use cautious radiology language when evidence is indirect.\\
7.~Output 3 to 5 complete sentences as one paragraph.\\[2.5pt]
\textbf{Structured answers}\\
\slotph{Agent C's question--answer pairs}\\[2.5pt]
\textbf{Mandatory incorporation rule}\\
The final caption must include one sentence about each of the following three liver
parenchymal attributes. A.~Diffuse parenchymal echogenicity. B.~Parenchymal echotexture,
coarse or heterogeneous. C.~Posterior acoustic attenuation. These three sentences must be
consistent with the structured answers, but must not mention the structured answers. If a
structured answer indicates not assessable, omit that attribute rather than guessing.\\[2.5pt]
\textbf{Prompt A (for reference)}~~\slotph{Agent A's instruction}\\
\textbf{Candidate A}~~\slotph{Agent A's caption}\\
\textbf{Prompt B (for reference)}~~\slotph{Agent B's instruction}\\
\textbf{Candidate B}~~\slotph{Agent B's report}\\[2.5pt]
\textbf{Task}\\
Write short, specific criticisms for Candidate A, Candidate B, and the structured
answers. Then write one final caption that preserves only image based and internally
consistent content.\\[2.5pt]
\textbf{Criticism restrictions}\\
1.~criticism\_A must critique only Candidate A text.\\
2.~criticism\_B must critique only Candidate B text.\\
3.~criticism\_C must critique only the structured answers quality or consistency.\\
4.~criticism\_A, criticism\_B, and criticism\_C must be different strings.\\
5.~Do not repeat the same sentence across criticism\_A/B/C.\\
6.~If one source is acceptable, still provide one concrete minor improvement specific to
that source.\\[2.5pt]
\textbf{Output format}\\
Return valid JSON with keys: rules\_check, verdict, criticism\_A, criticism\_B,
criticism\_C, judge\_final\_response.\\
rules\_check must be an object with keys R1, R2, R3 and brief pass or fail notes.
R1 checks forbidden non imaging content in the final caption. R2 checks diagnosis claims
in the final caption. R3 checks agent or structured answer mentions in the final caption.
verdict must be one short sentence describing which candidate is more reliable or whether
you mixed both. criticism\_A, criticism\_B, and criticism\_C must each be one short
concrete improvement point and must not be identical. judge\_final\_response must be the
final paragraph only, with 3 to 5 sentences, and must not mention agents or structured
answers.
\end{rolebox}
\caption{\textbf{The complete consensus-filtering judge prompt.} System and user messages
issued to the \SYS{} judge for every patient, verbatim from the inference code; only the
per-case content is replaced by \textcolor{vqablue}{$\langle$slots$\rangle$}. The three
rule families audited in \texttt{rules\_check} (R1: non-imaging content, R2: diagnosis
claims, R3: agent/VQA mentions) mirror hard constraints 1--3, and the mandatory
incorporation rule couples the caption to Agent C's three parenchymal attributes.}
\label{fig:prompt}
\end{figure}
\FloatBarrier

\suppnote{Model Capacity and Computational Cost}{note:compute}

\subsection{Trainable Adaptation Capacity}
\label{sec:capacity}

To examine whether greater trainable capacity explains the observed gains, we first
strengthen the single-VLM baselines used throughout the revised evaluation. MedGemma-4B
is increased to LoRA rank $r{=}32$, corresponding to $77.00$M trainable parameters, while
MedMO-4B uses $80.54$M trainable parameters. Both exceed the $55.63$M trainable budget of
\SYS{}, yet neither recovers its overall Dataset-1 performance in Tables~1--3 of the main
paper.

\begin{figure}[H]
\centering

\begin{minipage}[t]{0.48\textwidth}
    \centering
    \includegraphics[width=\linewidth]
    {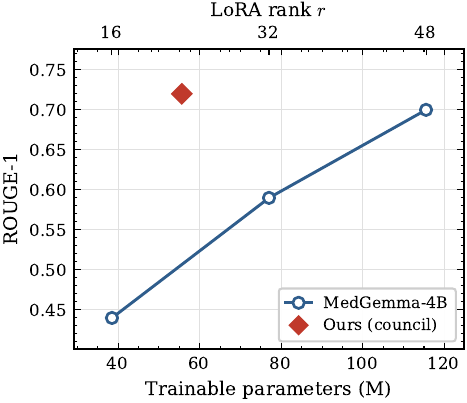}
    \par\vspace{1pt}{\small (a) ROUGE-1}
\end{minipage}
\hfill
\begin{minipage}[t]{0.48\textwidth}
    \centering
    \includegraphics[width=\linewidth]
    {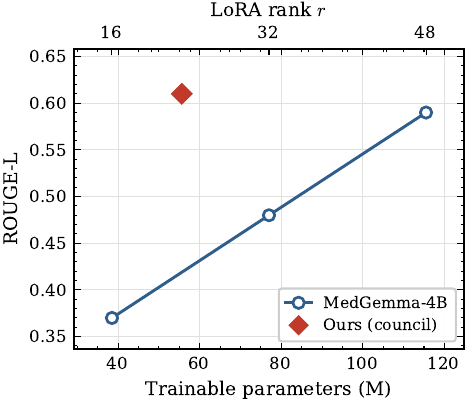}
    \par\vspace{1pt}{\small (b) ROUGE-L}
\end{minipage}

\vspace{5pt}

\begin{minipage}[t]{0.48\textwidth}
    \centering
    \includegraphics[width=\linewidth]
    {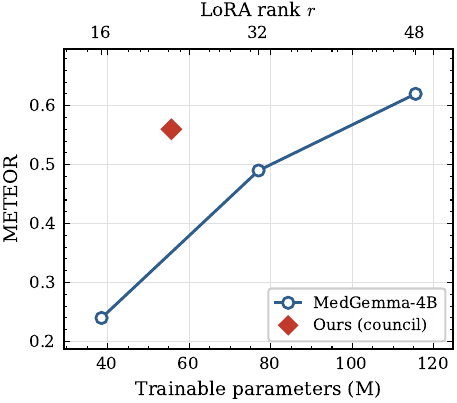}
    \par\vspace{1pt}{\small (c) METEOR}
\end{minipage}
\hfill
\begin{minipage}[t]{0.48\textwidth}
    \centering
    \includegraphics[width=\linewidth]
    {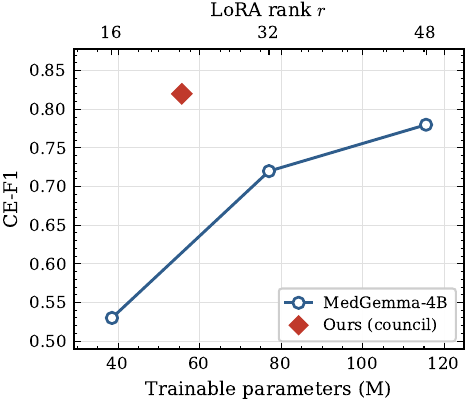}
    \par\vspace{1pt}{\small (d) CE-F1}
\end{minipage}

\caption{
\textbf{Effect of increasing single-model adaptation capacity.}
Performance of MedGemma-4B with increasing LoRA rank is compared with
SMILE on Dataset-1. Increasing the trainable capacity improves the
single-model baseline, while SMILE retains stronger ROUGE-1,
ROUGE-L, and CE-F1 performance at a lower trainable parameter budget.
}
\label{fig:capacity}
\end{figure}

We further analyze this trend by scaling MedGemma-4B over LoRA ranks
$r\in\{16,32,48\}$, corresponding to $38.50$M, $77.00$M, and $115.50$M trainable
parameters. Figure~\ref{fig:capacity} shows that increasing adaptation capacity improves
the single-model baseline consistently. From $r=16$ to $r=48$, ROUGE-1 increases from
$0.44$ to $0.70$, ROUGE-L from $0.37$ to $0.59$, and CE-F1 from $0.53$ to $0.78$.
However, the $r=48$ model uses approximately $2.1\times$ the trainable parameters of
\SYS{} while remaining below it on ROUGE-1 ($0.70$ vs.\ $0.72$), ROUGE-L
($0.59$ vs.\ $0.61$), and CE-F1 ($0.78$ vs.\ $0.82$).

\subsection{Computational Cost Accounting}
\label{sec:cost}

We use the generally used evaluation protocol for the computational-cost columns in the main tables. Every required model invocation is charged to the corresponding system.
Single-model baselines and MedAgents are allocated output budgets of $256$ and $512$ tokens per generative call, respectively. Dataset-1 contains $2$ views per patient ($512$ image tokens), whereas Dataset-2 contains an average of $3.96$ views per patient ($1015$ image tokens).

\begin{table}[H]
\centering
\footnotesize
\setlength{\tabcolsep}{4pt}
\begin{tabular}{@{}lrrr@{}}
\toprule
Component & Backbone & Rank & Trainable \\
\midrule
Agent A (radiology) & MedGemma-4B & $r{=}16$ & $38.50$M \\
Agent B (pathology) & Qwen2-VL-2B & $r{=}8$ & $14.61$M \\
Agent C (VQA)$^\ast$ & Qwen2-VL-2B & $r{=}8$ & $2.52$M \\
Judge & MedGemma-4B & frozen & $0.00$M \\
\midrule
\textbf{\SYS{} total} & & & $\mathbf{55.63}$\textbf{M} \\
\bottomrule
\end{tabular}
\caption{\textbf{Where \SYS{}'s trainable parameters sit.} The aggregation step is frozen, so extending the council with a new specialist requires
only the specialist adapter and no judge retraining. The parameters are
\emph{partitioned} across independently trained, individually reusable specialists: the
largest single training run ($38.50$M) is smaller than the MedGemma-4B ($77.00$M) and
MedMO-4B ($80.54$M) single-model baselines, each of which trains one $r{=}32$ adapter in a
single run. $^\ast$Agent~C adapts a restricted module set (query and key projections only),
which is why its budget is well below Agent~B's at the same rank.}
\label{tab:cost_supp}
\end{table}

\suppnote{Statistical Testing}{note:statistics}

\label{sec:stats}

For each NLG metric we compute per-patient scores for both systems on the same test
cohort and apply a two-sided Wilcoxon signed-rank test to the paired differences. We select a rank-based test because per-patient NLG scores are bounded and do not follow a normal distribution. We then correct for multiple comparisons across the family of tests within each table using the Holm--Bonferroni procedure. A dagger ($\dagger$) denotes $p_{\mathrm{adj}}<0.05$; unmarked NLG values are not statistically significant at that level.

\paragraph{Scope of inference.}
The Wilcoxon analysis quantifies paired output differences on the same patients and is not
used as an estimate of training-seed variability. CE-F1 is reported separately because the
main-table significance markers are restricted to patient-level NLG metrics. This scope is
stated explicitly in the revised Table~4 caption.

\FloatBarrier

\end{document}